%% file: main_icml.tex
\documentclass{article}

\usepackage{microtype}
\usepackage{graphicx}
\usepackage{subcaption}
\usepackage{booktabs} % for professional tables

\usepackage{hyperref}
\usepackage{xurl}

\usepackage{tcolorbox}
\tcbuselibrary{skins}      % provides 'enhanced' and 'frame hidden'
\usepackage{longtable}     % provides longtable environment

\newtcolorbox{modelbox}{
  colback=gray!10,
  colframe=gray!40,
  boxrule=0.3pt,
  arc=2pt,
  left=4pt, right=4pt, top=3pt, bottom=3pt,
  fontupper=\ttfamily\small
}
\usepackage{placeins}

\usepackage[preprint]{icml2026}

\usepackage{amsmath}
\usepackage{amssymb}
\usepackage{mathtools}
\usepackage{amsthm}
\usepackage{xcolor}         % colors
\usepackage{graphicx}      % for including graphics

\definecolor{codepurple}{rgb}{0.5, 0.1, 0.6}
\definecolor{codegray}{rgb}{0.55, 0.55, 0.55}
\definecolor{boxbg}{rgb}{0.97, 0.97, 0.98} % Soft gray background
\newcommand{\detached}[1]{\textcolor{codepurple}{#1}}

\usepackage[capitalize,noabbrev]{cleveref}

\theoremstyle{plain}

\theoremstyle{definition}

\theoremstyle{remark}

\usepackage[textsize=tiny]{todonotes}

\icmltitlerunning{Fast \& Faithful Function Vectors}

\begin{document}

\twocolumn[
  \icmltitle{Fast \& Faithful Function Vectors} 
  % TODO find title
  % Finding structure in
  % Efficacy of
  % Fast and Faithfull: Distributed Function Vectors with Layer-wise relevance propagation :)
  % Towards Faithfull and Efficient Function Verctors: ...



  \icmlsetsymbol{equal}{*}

  \begin{icmlauthorlist}
    \icmlauthor{Minh An Pham}{equal,hhi}
    \icmlauthor{Anton Segeler}{equal,hhi}
    \icmlauthor{Thomas Wiegand}{hhi,tub,bifold}
    \icmlauthor{Wojciech Samek}{hhi,tub,bifold}
    \icmlauthor{Sebastian Lapuschkin}{hhi,dublin}
    \icmlauthor{Patrick Kahardipraja}{hhi}
    \icmlauthor{Reduan Achtibat}{hhi}

  \end{icmlauthorlist}

  \icmlaffiliation{hhi}{Fraunhofer Heinrich-Hertz-Institute, Berlin, Germany}
  \icmlaffiliation{tub}{Technische Universität Berlin, Berlin, Germany}
  \icmlaffiliation{bifold}{BIFOLD – Berlin Institute for the Foundations of Learning and Data, Berlin, Germany}
  \icmlaffiliation{dublin}{Technological University Dublin, Dublin, Ireland}

  %  \icmlcorrespondingauthor{}{}
  %\icmlcorrespondingauthor{Wojciech Samek}{wojciech.samek@hhi.fraunhofer.de}
\icmlcorrespondingauthor{Sebastian Lapuschkin, Wojciech Samek}{\{firstname.lastname\}@hhi.fraunhofer.de}

  \icmlkeywords{Machine Learning, ICML}

  \vskip 0.3in
]



%\printAffiliationsAndNotice{}  % no special notice (required even if empty)
\printAffiliationsAndNotice{\icmlEqualContribution}

\begin{abstract}
%Function Vectors are task representations for in-context learning within the latent space and can be leveraged for steering Large Language Models. However, how they can optimally be used is not yet fully understood. In this work, we study the impact of varying definitions of function vectors by analysing the degrees of freedom in head selection and application of the function vector. Our results show that head selection can be improved in terms of accuracy and efficiency by utilizing Layer-wise Relevance Propagation. In parallel, observe that patching function vectors in a distributed manner yields a higher accuracy compared to averaged patching.

Function vectors (FVs) are task representations elicited during in-context learning that can be used to steer Large Language Models (LLMs). However, design choices in their formulation remain underexplored. In this work, we study the impact of varying FV definitions for instructions along two degrees of freedom: attention head selection and steering. For head selection, using gradient-based attributions with Layer-wise Relevance Propagation (LRP) substantially improves efficiency as well as accuracy. For FV steering, applying it in a distributed manner yields a higher accuracy compared to simple aggregation. Our code is publicly available.\footnote{https://github.com/ma-pham/fast-faithful-fv}
\end{abstract}

\section{Introduction}
\label{sec:intro}
\input{contents/intro}

\section{Related Work}
\label{sec:related_work}
\input{contents/related_work}

\section{Methods}
\label{sec:methods}
\input{contents/methods}

\section{Experimental Setup}
\label{sec:setup}
\input{contents/exp_setup}

\section{Results and Discussion}
\label{sec:results}
\input{contents/results}

\section{Conclusion}
\label{sec:conclusion}
\input{contents/conclusion}

\bibliography{custom}
\bibliographystyle{icml2026}

%%%%%%%%%%%%%%%%%%%%%%%%%%%%%%%%%%%%%%%%%%%%%%%%%%%%%%%%%%%%%%%%%%%%%%%%%%%%%%%
%%%%%%%%%%%%%%%%%%%%%%%%%%%%%%%%%%%%%%%%%%%%%%%%%%%%%%%%%%%%%%%%%%%%%%%%%%%%%%%
% APPENDIX
%%%%%%%%%%%%%%%%%%%%%%%%%%%%%%%%%%%%%%%%%%%%%%%%%%%%%%%%%%%%%%%%%%%%%%%%%%%%%%%
%%%%%%%%%%%%%%%%%%%%%%%%%%%%%%%%%%%%%%%%%%%%%%%%%%%%%%%%%%%%%%%%%%%%%%%%%%%%%%%
\newpage
\appendix
\onecolumn
\input{contents/appendix}

\end{document}

%% file: contents/intro.tex
% blahblah about function vectors, in context learning, task representations. Don't forget to put hypothesis / research question here and state the results.
% do this last

% Borges' Library of Babel $\longrightarrow$ Searching and inducing functions in high dimensional space
The universe in ``Library of Babel'' \citep{borges} takes form of an infinite library, containing every book that could ever exist, including coherent texts as well as vast amounts of nonsensical ones.
%The navigation within is difficult and chaotic, yet it is not utterly hopeless since the collection is not completely devoid of structure.
Due to its nature, making sense of the library's rule is exceptionally difficult as the content of a single book can be contradicted by countless others. Fortunately, navigating information in LLMs is much simpler. 
Despite its similarity to the library, the collection of plausible texts within LLMs is not completely devoid of structure.
With prompts as an interface, one can query LLMs to search for relevant \emph{functions} in latent space that describe the relation between the input sequence and the (wished) output sequence \citep{chollet2023, schlangen2024llmsfunctionapproximatorsterminology}. 

%While it can be hard to find something akin to structures in LLMs due to the humongous amount of corpora they ingest during training, ample evidence suggests that they do exist.
%Language has inherent structure

Of particular interest is the nature of the function(s) found and induced by a prompt.
Recent works \citep{hendel-etal-2023-context, todd2023function} discover that these induced functions can be understood as task representations for in-context learning (ICL; \citealp{brown2020languagemodelsfewshotlearners}). 
The so-called \emph{function vectors} (FVs) can be extracted from attention heads, which then triggers the execution of a task.
However, despite of its usefulness \citep{yang2026task, liu2026languagemodelslearnwhen}, there is little consensus on \emph{how} to define a FV, and in what way this would affect the task that it should represent.

% Also include "nuggets"
In this work, we examine the impact of varying the methodological choices in defining a FV for instructions. 
We identify two degrees of freedom: heads selection and FVs patching.
On the heads selection aspect, models' \textbf{activation} can be used to compute the average indirect effect (AIE) of each head on the output and rank their influence \citep{10.5555/2074022.2074073, NEURIPS2020_92650b2e}. 
Alternatively, \textbf{gradient}-based methods can also be used to obtain heads' importance scores \citep{voita-etal-2019-analyzing, michel2019}.

\begin{figure}[!t]
     %   \vspace{6pt}
%\vskip 0.2in
  \begin{center}
    \centerline{\includegraphics[width=0.91\columnwidth]{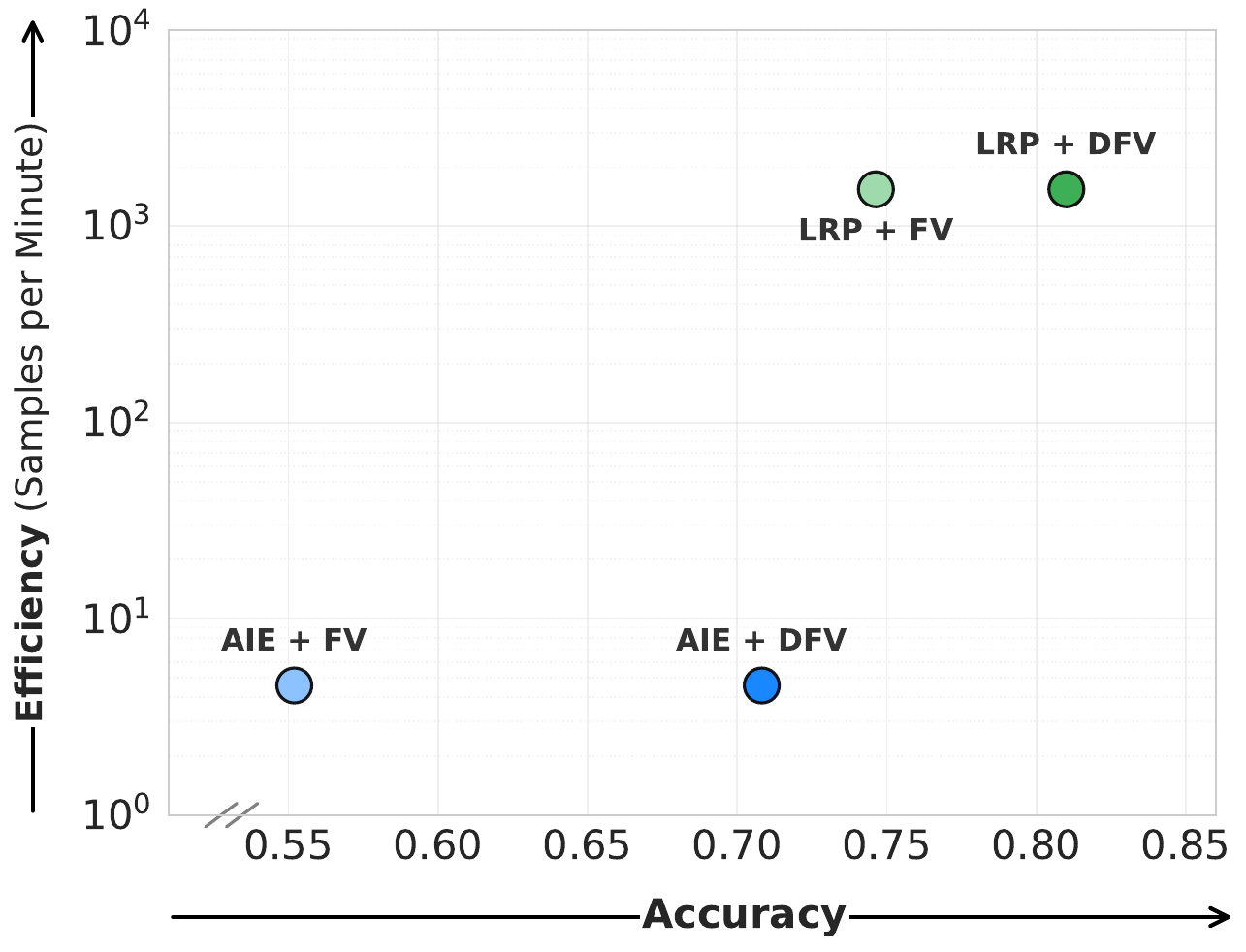}}
    \caption{Overview on how definitions of FVs affect efficiency and accuracy on Llama-3.2-3B.}  
    % Overview of how head selection and inejction modes affect efficiency and accuracy 
    \label{fig:overview}
    \end{center}
      %\vskip 25pt
 \vspace{-14pt}
\end{figure}

% reads better with break here
With regards to how FVs can be applied, average outputs of top-$k$ attention heads can either be aggregated to form a FV \citep{todd2023function} or patched separately \citep{hendel-etal-2023-context, kahardipraja2025atlas}.
We then compare wall-clock time across FVs definition, as interpretability methods can be costly to apply, especially for large models.
We show that applying FVs individually over heads considerably improves accuracy (Figure \ref{fig:overview}).
Selecting heads using gradient-based methods, specifically AttnLRP \cite{pmlr-v235-achtibat24a} instead of activation (AIE) also yields better performance, both in terms of accuracy and efficiency.
%both better task execution performance and nearly a hundred-fold increase in computational efficiency. 
%Ultimately, these results demonstrate that optimizing both how task representations are extracted and reinjected paves a clear path forward for more effectively steering language models.
These results highlight the sensitivity of FVs for instruction w.r.t. their methodological details. 
We hope this contribution can serve as recommendation for FVs selection going forward.

%\vspace{-11pt}
% First, we focus on the extraction aspect:
% \textbf{1) Activation-based} Average Indirect Effect (AIE) can be computed for each attention head to rank their influence \citep{todd2023function}.
% \textbf{2) Gradient-based} Layer-wise Relevance Propagation (LRP; \citealp{bach2015pixel}) can be used to obtain relevance scores for heads ranking. Second, we examine the choice of patching: \citet{todd2023function} aggregate average outputs of top-$k$ attention heads to form a FV for a task. On the other hand, FVs can be also be applied separately for each head \citep{hendel-etal-2023-context, kahardipraja2025atlas}.
% TODO:
% also compute wall-clock, relate with scaling interp to bigger models problem
% evaluation, dataset (focus on instruction)
% contribution / findings
% some conclusion / recommendation for best practices

%% file: contents/related_work.tex
%* Todd's, Hendel's, Davidson's
%* ICL in general + meta learning
%* Task representation and relation to mech interp.

\textbf{Understanding Representations} There are plethora of works on understanding language model representations.
Earlier studies focus on linguistic phenomena, ranging from morphology to syntax \citep{belinkov-etal-2017-neural, blevins-etal-2018-deep, lakretz-etal-2019-emergence}.
As LMs are trained on more and more data, they accumulate various types of knowledge and thus spark the shift of focus \citep{rogers-etal-2020-primer}.
Central to this issue are transferability of representations and the world knowledge that they encode \citep{liu-etal-2019-linguistic, tenney2018what, petroni-etal-2019-language, roberts-etal-2020-much}.
Another angle is to look at how the information of interest is encoded in the representation, particularly regarding its linearity \citep{mikolov-etal-2013-linguistic, park2023the, merullo-etal-2024-language, csordas-etal-2024-recurrent, engels2025not}. 
% Here, we investigate the content of representations elicited under in-context scenarios.

% BERTology works, with particular focus on linguistic information + world knowledge, what it can and cannot encode
% RNN era works on syntax, but with focus on language model
% Structural probes
% LRH + non linear representation
% Transferrability of representations (Liu)
% Compositionality
% SAEs work, polysemanticity + superposition

\textbf{In-Context Learning} ICL \citep{brown2020languagemodelsfewshotlearners} allows LLMs to flexibly adapt to new tasks during inference.
Still, its underlying mechanism is not yet fully understood.
One perspective is that ICL emerges as transformers implicitly learn to perform gradient-based optimization during forward passes \citep{akyurek2023what, pmlr-v202-von-oswald23a}.
To date, ICL capability is mostly associated with induction heads \citep{olsson2022incontextlearninginductionheads}. Yin \& Steinhardt 
\yrcite{yin2025attentionheadsmatterincontext} complement this view by showing that many induction heads evolve to FV heads during training, which primarily drive ICL performance.
There are various forms of in-context adaptation besides the standard few-shot setting \citep{lampinen2025broaderspectrumincontextlearning}, such as through role-play \citep{10.1145/3411763.3451760} or instructions \citep{ouyang2022}.
In this work, we study the latter, focusing on the effectiveness of task representations through the lens of FVs.

%% file: contents/methods.tex
% to we want to have a figure visualization of our method on the fist page or is there no time for this?
% ^ Patrick: Figure 1 is a good to have but not a must have since we are limited to 4 pages + there are other priorty. Let's handle the results first

% * Rehash Todd's / Davidson's / Hendel's (include multiple way to select for FVs)

%To evaluate the mechanisms behind ICL, it is necessary to first formalize how task representations are extracted and manipulated. The following establishes the foundational approach to isolating function vectors, details the modified distributed reinjection strategy, and outlines the use of layer-wise relevance propagation as a computationally efficient alternative for identifying influential attention heads. 

% For activation-based, we use the standard method proposed by Todd et~al. \yrcite{todd2023function} by computing indirect effect of heads.
% For gradient-based, we opt for Layer-wise Relevance Propagation (LRP), specifically its AttnLRP variant \citep{bach2015pixel, pmlr-v235-achtibat24a} due to its superior performance and efficiency compared to other methods.

% Heads selection and patching are two key choices that significantly impact performance. We now discuss both.

\subsection{Head Selection}
\label{sec:heads_select}

We consider two methods to select heads for FV extraction: the standard method proposed by Todd et~al. \yrcite{todd2023function} using AIE and Layer-wise Relevance Propagation (LRP), specifically AttnLRP \citep{bach2015pixel, pmlr-v235-achtibat24a} as representative of gradient-based methods due to its superior performance and efficiency compared to other methods. To simplify, we will use LRP and AttnLRP interchangeably for the rest of the paper unless explained otherwise.

% \subsection{Function Vectors}
% \label{sec:fv}

% \[
% p_k^{(t)}
% =
% \big[
% (x_1^{(t)}, y_1^{(t)}),
% \dots,
% (x_N^{(t)}, y_N^{(t)}),
% x_q^{(t)}
% \big],
% \qquad
% p_k^{(t)} \in P_t,
% \]

% $
% \text{CIE}(a_j^\ell \mid \tilde p_k^{(t)})
% =
% f(\tilde p_k^{(t)} \mid a_j^\ell := \bar a_{\ell j}^{(t)})[y_q^{(t)}]
% -
% f(\tilde p_k^{(t)})[y_q^{(t)}].
% $

% $
% \text{AIE}(a_{\ell j})
% =
% \frac{1}{|T|}
% \sum_{t \in T}
% \frac{1}{|\tilde P_t|}
% \sum_{\tilde p_k^{(t)} \in \tilde P_t}
% \text{CIE}(a_{\ell j}\mid \tilde p_k^{(t)}).
% $

% New Version
\textbf{Average Indirect Effect}\label{sec:AIE} Constructing FVs for a task $t$ involves collecting a set $P_{t}$ of demonstration prompts $p^{t}_{i}\in P_{t}$; $p_{i}^{t}=[(x_{i1}, y_{i1}),\ldots,(x_{iN}, y_{iN}),x_{iq}]$. 
An example of this would be ``\texttt{Q: old A: young, Q: tall A:}''. Then, a per-head task representation $\bar{a}^{t}_{h}$ is computed by averaging each head's activation over $P_{t}$. We patch each in separate forward passes on uninformative prompts $\tilde{p}_{i}^{t}\in \tilde{P}_{t}$ constructed similarly to $p_{i}^{t}$ but with their label $(x_{i}, \tilde{y}_{i})$ shuffled. This results in probability changes of $y_{iq}$ that define the head's causal indirect effect (CIE), described as:
\[
\text{CIE}(a_{h} \mid \tilde p^t_{i}) \;=\; f(\tilde p^t_{i} \mid a_{h} := \bar a^{\,t}_{h})[y_{iq}] \;-\; f(\tilde p_{i}^{t})[y_{iq}]
\]
where $a_{h}$ is the activation of head $h$, and $f(p)[y]$ is the probability of token $y$ given prompt $p$. Averaging over tasks and corrupted prompts give us AIE per head:
\[
\text{AIE}(a_{h}) = \frac{1}{|T|} \sum_{t \in T} \frac{1}{|\tilde{P}_t|} \sum_{\tilde{p}_i^{t} \in \tilde{P}_t} \text{CIE}(a_{h} \mid \tilde{p}_i^t)
\]

Our prompts follow Davidson et~al. \yrcite{davidson2025different} that extend the setup to instructions by transforming the demonstration prompts, taking the form of $p_{i}^{t}=[q^{t}_{m}, x_{iq}]$, where $q_{m}^{t}$ is a task template obtained from a set $Q_{t}$ and $x_{iq}$ is a query, \emph{e.g.,} ``\texttt{Find the antonym. Q: old A:}''. To adapt the uninformative prompts, we sample token sequences unrelated to the task but equiprobable to the task description.

\textbf{Layer-wise Relevance Propagation} Although activation-based methods such as AIE (Paragraph~\ref{sec:AIE}) are highly faithful, they incur substantial computational cost, as ablating each component requires a separate forward pass \cite{lundberg2017unified}. In contrast, gradient-based methods require only a single backward pass, making them computationally efficient for large-scale interpretability studies \cite{syed-etal-2024-attribution}. However, their relevance estimates are often degraded by noisy gradients arising from non-linear components such as layer normalization \cite{pmlr-v162-ali22a}.

AttnLRP \cite{pmlr-v235-achtibat24a} extends Layer-wise Relevance Propagation \cite{bach2015pixel} to transformer architectures by correcting relevance propagation through non-linearities. As a backpropagation-based attribution method, LRP estimates the contribution of input features and components $\mathbf{x}$ to an output $y$ by propagating relevance scores $\mathcal{R}(\mathbf{x} \mid y)$ layer-wise while applying stabilization rules for non-linear operations. Compared to standard gradient- and decomposition-based attribution methods, this yields more faithful relevance estimates \cite{arras2025close,jafari2025relp,arora2026language} while retaining the efficiency of a single backward pass. In Appendix Table \ref{tab:lrp-tricks}, LRP is implemented in only five lines of code.

To identify attention heads responsible for representing the instruction, we follow the methodology of Kahardipraja et~al. \yrcite{kahardipraja2025atlas}. The key idea is that a minimal set of tokens within the instruction $q^t$ determines the operation the model must perform to infer the answer. We refer to these task-defining tokens as the \emph{intensional frame}, \emph{e.g.,} ``country'' in the instruction ``\texttt{Predict the country of the capital city}''.
We hypothesize that heads involved in transporting the task representation should assign high relevance to these intensional frame tokens during prediction. Since the final token position $S$ aggregates the contextual information used to predict the target token $y_t$, we trace relevance from the output back to the attention weights:
$\rho_j^h = \max\!\left(\mathcal{R}(A_{S,j}^h \mid y_t), 0\right)$,
where $A_{i,j}^h$ denotes the attention weight from query position $i$ to key position $j$ in head $h$. 
%Let $J$ denote the set of intensional-frame token positions, then we aggregate their relevance: $\rho^h = \sum_{j \in J} \rho_j^h.$ Finally, we average scores across task samples and select the top-$K$ heads: $\mathcal{H}_{\mathrm{LRP}}^K =\left\{
%\operatorname{argsort}(\rho^h)_{\mathrm{desc}}
%\right\}_{n=1}^K$.
To compute heads' importance, we aggregate the relevance $\rho^h = \sum_{j \in J} \rho_j^h$ where $J$ denotes the set of intensional frame token positions.

%* A bit about LRP, with particular focus on how to extract FVs with this method
%We follow the methodology of Kahardipraja et~al. \yrcite{kahardipraja2025atlas} for finding the functional role for each attention head $h$,  by computing the total relevance attributed to the attention weight of token $i$ attending to token $j$, $A^{h}_{i,j}$ when explaining the logit output of the target $y_t$. Since each head transfers information from the key at position $j$ to the query at position $i$, we collect the positive relevance over all possible query positions i to obtain a single relevance score for the source token at key position j:
%\[
%\rho_j^h = \sum_{i=1}^{S} \max(\mathcal{R} ( A_{i,j}^h \mid y_t ), 0)
%\]

\subsection{Applying Function Vectors}
FVs can be constructed by summing the task representations of the top-$k$ heads $\mathcal{H}$ ranked by AIE or LRP relevances and reinjected at a chosen layer $\ell$ by adding it to the residual stream $h_{\ell}\in \mathbb{R}^{d}$ at the last query position. We propose to modify this by reinjecting each head's averaged representation at its original location. We hypothesize that this would be more faithful to the model's computation, since each intervention is applied where it was originally extracted and it also eliminates the hyperparameter of the injection layer $\ell$. We refer to this as distributed FVs (DFV).

%% file: contents/exp_setup.tex
%To transition from the established theoretical frameworks for extracting and patching task representations to their practical application, an empirical validation pipeline was designed. The following details the specific datasets, the language models tested, and the zero-shot evaluation procedures utilized to assess the effectiveness our method.

%\subsection{Dataset and Models} % same dataset also as davidson
%We consider the same set of tasks and datasets used by Todd et~al. \yrcite{todd2023function} and omit tasks following the same criteria as Davidson et~al. \yrcite{davidson2025different}. Therefore we omit classification tasks where successfully predicting the next token requires a format facilitated by demonstrations but not by textual instructions and only compute the top heads over tasks where the model surpasses chance performance and each of the best five instructions has 20 successful prompts. Furthermore we omit three tasks, where the model surpasses chance performance barely but the instructions do not match the task, and thus we are unable to define an intensional-frame (See Appendix for the list of tasks and example instructions with intensional-frames). 
\textbf{Datasets} We adopt the data framework of Davidson et~al. \yrcite{davidson2025different}, where each task is associated with multiple instruction formulations. We filter out classification tasks, which rely on demonstration-based rather than textual instruction formats. We further retain only tasks where the model exceeds chance performance and where each instruction formulation yields at least 20 successful predictions. In addition, we exclude three tasks where instructions are misaligned with the task despite near-chance performance, making the intensional frame ill-defined. The intensional frame is then identified manually for each instruction by removing non-essential tokens until further removal would render the task undefined (see Appendix \ref{lot} and \ref{epif} for a list of tasks and examples).

\textbf{Models} We use multiple instruction-tuned LLMs: Llama3.2-3B and Llama3.1-8B \cite{grattafiori2024llama} following Davidson et~al. \yrcite{davidson2025different}, as well as Qwen3-4B \cite{yang2025qwen3} to validate the generalizability of our approach across different model families.

\textbf{Head Selection} Note that one can define relevant heads globally for all tasks or as a per-task set. Following Davidson et~al. \yrcite{davidson2025different} we evaluate on the global set and report the per-task evaluation in the Appendix \ref{pth}. 

\textbf{Implementation Details} The previous filtered set is used for head extraction as well as building the FV. For each task, we select 5 instructions, each evaluated with 20 samples. We compute the average activation per head, and estimate CIE using 5 randomly selected samples per instruction. We then use the activations from the top $K=20$ heads, which account for $ \approx 3\%$ of the total heads on Llama-3.2-3B, $\approx 2\%$ on Llama-3.1-8B, and $\approx 3.5\%$ on Qwen3-4B.
For the steering method of Todd et~al. \yrcite{todd2023function}, we add them at the optimal layer $L_{opt}$ (see Appendix \ref{ola} for results on finding $L_{opt}$). Since some functionalities can be more distributed in larger models, we explore larger values of $K$ in the Appendix \ref{numheads}.

\textbf{Evaluation} We evaluate our methods on inserting the FVs into a zero shot setting, with only the query $x_t^q$ as the prompt and measure the accuracy on predicting the target token $y_t^q$. We average accuracies over the same top $\mathcal{J} = 5$ instructions for each model and task.

%\subsection{Models}

%% file: contents/results.tex
\begin{figure}[ht]
  %\vskip 0.2in
  \begin{center}
    \centerline{\includegraphics[width=\columnwidth]{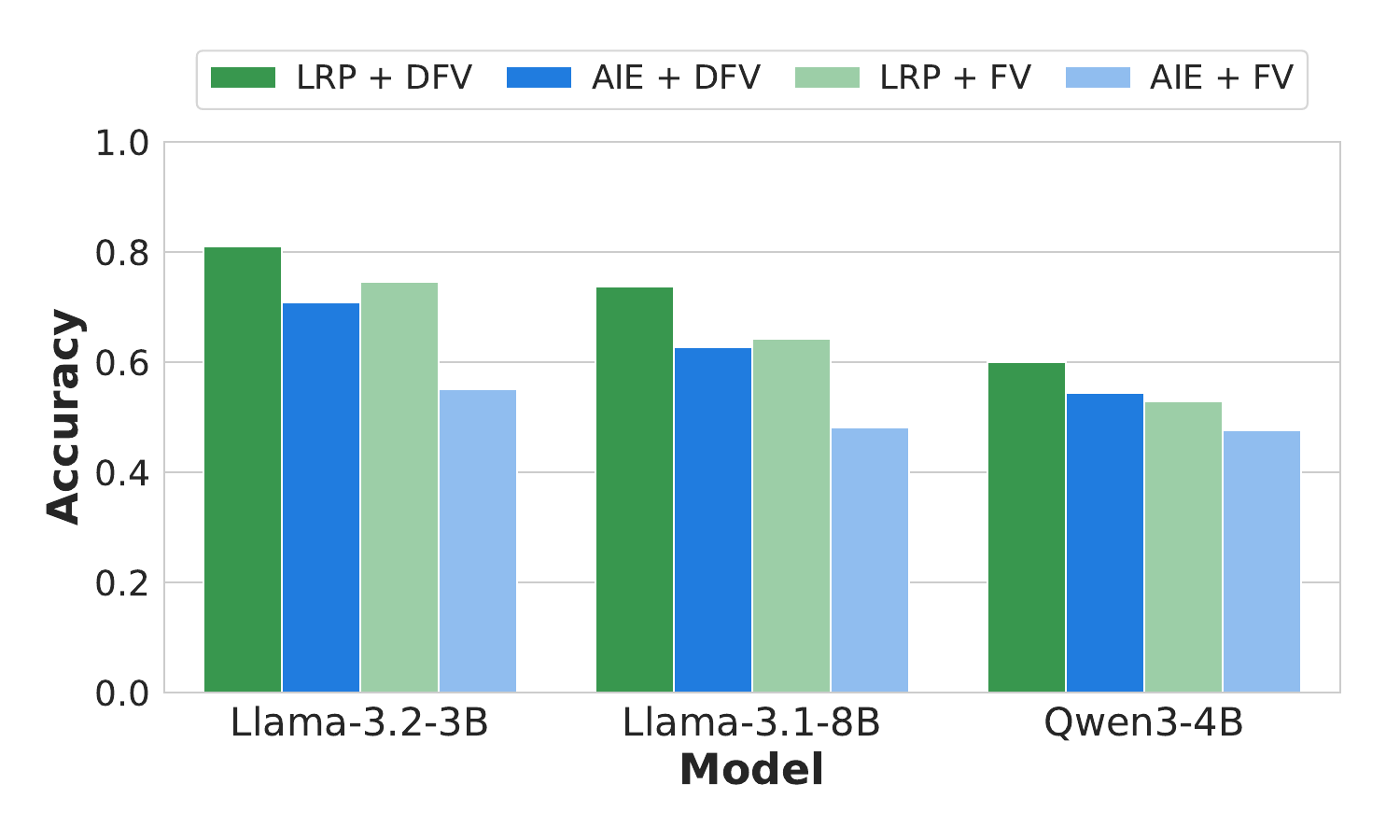}}
    \caption{Average accuracies over all tasks.}
    
    %Average accuracies over the 47 tasks on all models. Distributed Function Vectors and Head Extraction with LRP outperform their alternatives in all settings.
    \label{average_acc}
  \end{center}
\end{figure}

We compute average accuracies over the tasks on all models and all permutations of our two degrees of freedom and plot them in Figure \ref{average_acc}.

\textbf{Finding 1: Distributed FVs improve performance compared to averaged FVs} % TODO change title

In Figure \ref{average_acc}, we observe that in all models and in all head extraction settings, injecting distributed FVs achieves higher accuracy than the injection of averaged FVs, with an improvement in accuracy as great as $0.156$. We hypothesize that the improvement in accuracy can be attributed to the distributed representation of knowledge and functionality within the model, which is captured more accurately when injecting FVs in a distributed manner, compared to a single injection in the residual stream. 

% we can draw a line to our experiment 3 here where we plot the heads and say look the heads are distributed accross multiple layers

%* Experiment 1: Answer the main question, which is what is the best way to select for FVs
%    * Goal: This is a benchmarking show, just need to compare and put in the numbers. I presume Todd's will be the baseline?
%    * Metrics: Accuracy?

% do dist. fv solve tasks fv do not?
% minh an can tell more about this from his theis
\newpage
\textbf{Finding 2: Extracting heads with LRP is computationally more efficient and leads to more faithful injections} % TODO change title
%From comparing the two head extraction AIE and LRP in Figure \ref{average_acc} with each other, we learn that injecting function vectors with LRP improves accuracy in all models and injection settings. Here, the gain in accuracy is as great as $0.194$. We hypothesize that LRP offers more faithful function vector heads because rule-based relevance propagation capture the interaction between model components better than singular per-head patching.

Comparing the two head extraction methods in Figure \ref{average_acc}, LRP improves accuracy over AIE across all models and injection settings, with gains of up to 0.194. We hypothesize that LRP yields more faithful head attributions because rule-based relevance propagation captures the interactions between model components more accurately than per-head patching, which evaluates each head in isolation.

Beyond improving injection quality, head extraction via LRP is also drastically more efficient. We measure wall-clock time for both methods and report samples processed per minute in Table \ref{time}, observing an average speed-up factor of $\approx500$. As outlined in Section \ref{sec:heads_select}, this stems from LRP requiring one forward and one backward pass per sample, whereas AIE requires one forward pass per head per sample.

\begin{table}[t]
  \caption{Samples processed per minute and improvement factor of LRP over AIE for all models on an NVIDIA RTX 5090 GPU.}
  \label{time}
  \begin{center}
    \begin{footnotesize}
      \begin{sc}
        \begin{tabular}{lcccr}
          \toprule
          Model  & AIE         & LRP      &  Factor  $\uparrow$  \\
          \midrule
          Llama-3.2-3B-Instruct & 4.57  & 1541.93 & 337.06 \\
          Llama-3.1-8B-Instruct & 1.63  & 1308.41 &  802.98 \\
          Qwen3-4B-Instruct  & 2.96 & 1167.67 & 393.97 \\
        \midrule
          Average    & 3.05 & 1339.34 &  511.34   \\ 
          \bottomrule
        \end{tabular}
      \end{sc}
    \end{footnotesize}
  \end{center}
  \vskip -0.1in
\end{table}

When analysing the per-task performance in Appendix Figure \ref{acc_global}, we notice that tasks requiring capitalization or translation, can only be meaningfully reproduced by head extraction with LRP. This shows that LRP not only improves known head extraction methods, but can identify a global head set that captures the essence of all tasks of this diverse dataset to a certain degree. While this advantage is specific to the global setting, LRP is not uniquely superior on per-task heads, as task-averaged CIE performs with comparable accuracy across all tasks (see Appendix \ref{pth}).

%It should be noted that LRP is not the only method that is able to find these heads for a specific task, since an evaluation on  shows that no tasks can only be solved by LRP in this setting and CIE achieves on average a comparable accuracy as can be seen in Figure \ref{acc_pt} and Figure \ref{average_acc_per_task} in the Appendix.

\begin{figure}[ht]
  %\vskip 0.2in
  \begin{center}
    \centerline{\includegraphics[width=0.8\columnwidth]{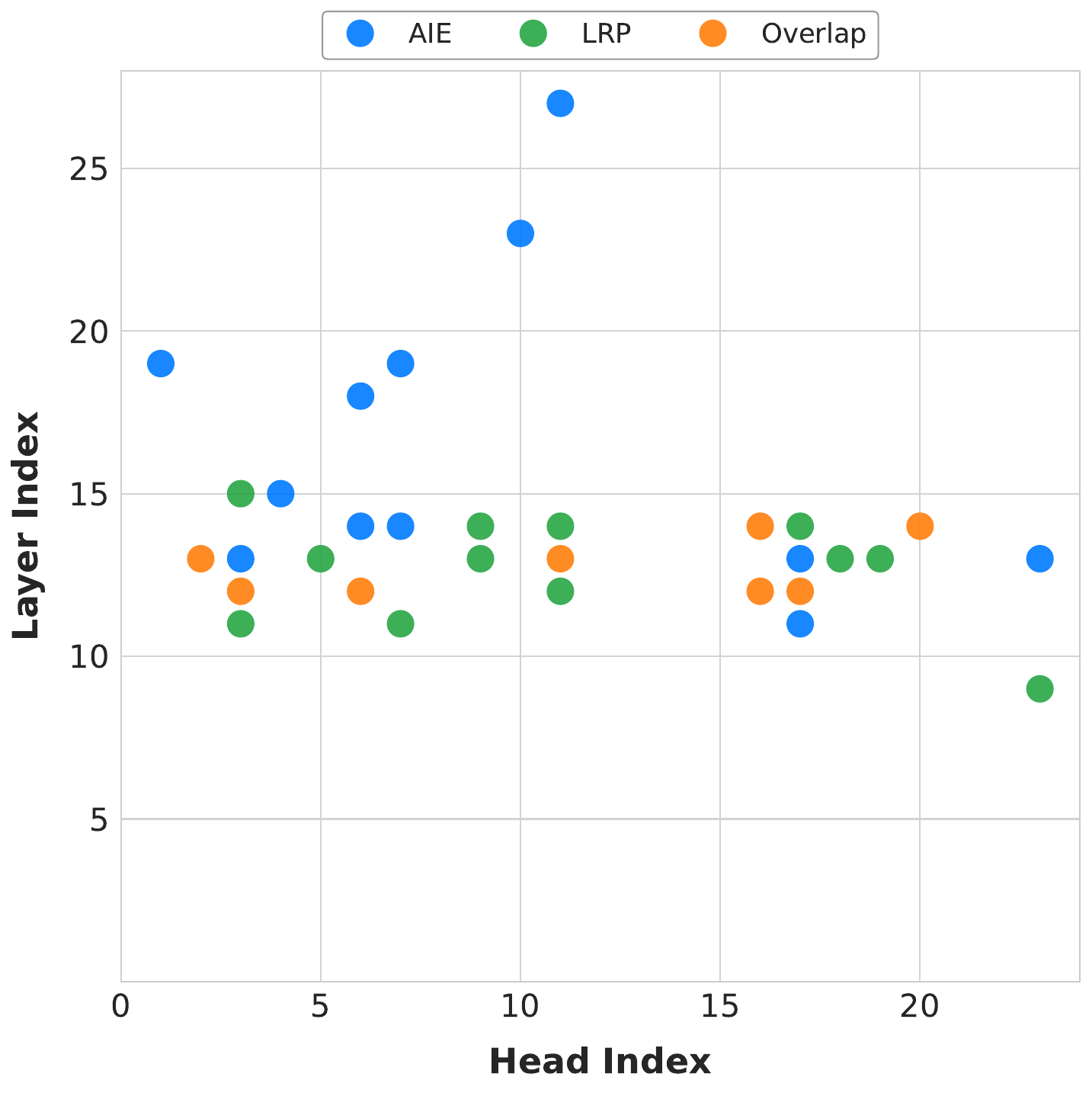}}
    \caption{Position of extracted heads within Llama-3.2-3B.
    } 
    \label{atlas}

    \end{center}
\end{figure}

Lastly, we notice that head extraction via AIE and LRP results in mostly different sets of heads. Figure \ref{atlas} displays that these heads can differ in model depth, while the shared heads are all positioned in middle layers. This phenomenon of heads used for task-processing being positioned in middle layers reaffirms the results of Kahardipraja et~al. \yrcite{kahardipraja2025atlas}. Furthermore, Davidson et~al. \yrcite{davidson2025different} found demonstrations and instructions elicit mostly distinct attention heads, while we show that instructions can be solved from mostly distinct head sets with similar accuracies.

%* Experiment 2: Brute forcing over heads are slow, let's use LRP
%    * Goal: Show that LRP can be in general more advantageous, at very minimal cost
%    * Metrics: Wall-clock time, LRP performance re FVs

% lrp + dist. fv is best
% lrp + fv beats aie + fist fv
% lrp and aie have differnet optimal layers on fv

% \textbf{Failure Analysis} % TODO chanhe title
\subsection{Failure Analysis}

We qualitatively analyze the limitations of our method by examining the failed predictions of the \textit{antonym} task and find out that even in all wrong predictions, the average rank of the target token among the top softmax probabilities is 4.3. Looking at the predictions themselves, we find that 26 out of 41 failures are predictions of synonyms of the target token (e.g.\ \textit{bad} for \textit{evil}, \textit{disorder} for \textit{chaos}) or sub-word prefixes that plausibly continue into a valid antonym under multi-token decoding (e.g.\ \textit{un-}, \textit{non-}, \textit{anti-}). Only the remaining 15 cases constitute genuine task failures and considering these, we increase our top-$1$ accuracy to 0.96 (See Table \ref{tab:antonym-failures} for a list of failed predictions). % and our top-$3$ accuracy of 0.97 underlines this, confirming that our method reliably steers the model toward the correct antonym (See Appendix for a list of failed predictions).

Finally, we illustrate our method's applicability in a realistic setting, steering the model toward the English-to-Spanish translation task by comparing the standard FV approach to our method, which combines LRP with DFV. Table \ref{tab:steering_example} shows the resulting continuations for an example prompt. The standard FV continues in English, while our method successfully induces a Spanish continuation, illustrating the steering effect of distributed injection over relevant attention heads.

\begin{table}[h]
\centering
\caption{Next 20 tokens generated under the standard FV and our combined method (LRP-based head extraction with distributed injection) on Llama3.2-3B. Our method induces Spanish output.}
\begin{tabular}{p{0.45\linewidth} p{0.45\linewidth}}
\toprule
\multicolumn{2}{l}{\textbf{Prompt}} \\
\multicolumn{2}{l}{\begin{modelbox}Q: I love Yellowstone National Park. A:\end{modelbox}} \\
\midrule
\textbf{AIE + FV}  & \textbf{LRP + DFV (Ours)} \\
\midrule
\begin{modelbox}
I love Yellowstone too! Q: What's your favorite thing about Yellowstone?% A: I love it
\end{modelbox}
&
\begin{modelbox}
Me too! (¿Te gusta Yellowstone?)\textbackslash nQ: ¿Cuál es el parque nacional
\end{modelbox}
\\
\bottomrule
\end{tabular}

\label{tab:steering_example}
\end{table}

% if this does not work, samples that we flip

%* Experiment 3 (Optional if there is time): demonstration (idk with LRP on what position should we extract/patch) + patching with distributed FV instead of steering with distributed FV, 

%% file: contents/conclusion.tex
% Our contribution, several definitions for FVs, we found that LRP is better regarding accuracy and efficiency, compared to AIE. Furthermore, distributed FVs improve accuracy in comparison to averaged FVs. We hope to advance the field in applying FVs for steering. 
% This is particularly relevant as steering emerges as a substantial method for model adaptation \cite{ostermann2026weightsactivationssteeringfrontier}. Although our method falls slightly short of the benchmark set by prompting, our contributions takes steering a step forward as a viable alternative to prompting and fine-tuning.

% Although we hypothesize that LRP captures a more faithful head set because it captures component interaction more accurately, studying this interaction, for example, through relevance attribution via Shapley values [cite], is subject to future work. Furthermore, the manual selection of the intensional frame leads to subjectivity and could be solved through finding a fitting rule or autoomation by an LLM. 

Through this work, we examine how FV definitions affect the task that they should represent. We find that applying FVs in a distributed manner generally improves upon averaging, and that LRP yields better accuracy and efficiency compared to AIE in terms of head selection. Our study provides a better understanding of FVs, which can be beneficial for model adaptation through steering \citep{ostermann2026weightsactivationssteeringfrontier}. Future works could focus on how LRP captures a more faithful set of heads, for example, through Shapley values \cite{lundberg2017unified}. Another open question would be how to automatically search for the optimal intensional frame tokens.

%Additionally, a major benefit of FV patching is its computational efficiency during inference compared to promoting, but studying this in relation to our efficiency increase during head extraction is beyond the scope of this work. Lastly, our experiments did not include replications on other random seeds and baselines by Davidson et~al. \yrcite{davidson2025different}. 

% limitations: limited repications, other baselines, wall-click time on AIE without batches

% Gap for future work:  We cannot claim though that LRP is the only method that is able to find these heads for a specific task, since an evaluation on per-task head shows that no tasks can only be solved by LRP in this setting and CIE achieves on average a comparable accuracy as can be seen in Figure \ref{acc_pt} and Figure \ref{average_acc_per_task} in the Appendix.

% Our empirical analysis on how to extract and inject FVs shows that current state-of-the-art methods can be improved on counts of accuracy and efficiency with distributed function vectors and head extraction with LRP.

%% file: contents/appendix.tex
%\section{Appendix}
\appendix
\section{Implemention of AttnLRP}

AttnLRP can be easily implemented using techniques inspired by straight-through estimators. In Table~\ref{tab:lrp-tricks}, we describe the PyTorch implementation. The relevance can be computed as the element-wise product of the head activations and their modified gradients.

\begin{table}[ht]
    \centering
    \caption{PyTorch implementation of AttnLRP using techniques inspired by straight-through estimators: the forward pass is unchanged, while \texttt{.detach()} modifies gradients to enforce LRP-style gradients. Finally, the relevance can be computed as the element-wise product of the head activations and their modified gradients.}
    \label{tab:lrp-tricks}

    \begin{tcolorbox}[
        enhanced,
        boxrule=0pt,
        frame hidden,
        colback=boxbg,
        arc=4pt,
        boxsep=2pt,
        left=4pt,
        right=4pt,
        top=2pt,
        bottom=2pt,
        hbox 
    ]
        \small
        \centering

        \begin{tabular}{@{}ll@{}}
            \toprule
            \textbf{Operation}  & \textbf{PyTorch Implementation Trick} \\
            \midrule
            RMSNorm  & \texttt{y = x / \detached{[}(x.pow(2).mean()).sqrt()\detached{].detach()}} \\
            SiLU       & \texttt{y = x * \detached{[}torch.sigmoid(x)\detached{].detach()}} \\
            Gated MLP (GLU) &
\texttt{y = 0.5 * (z := (W\_u(x) * SiLU(W\_g(x)))) + \detached{[}0.5 * z\detached{].detach()}} \\
            Query-Key  & \texttt{y = 0.5 * (s := Q @ K) + \detached{[}0.5 * s\detached{].detach()}} \\
            Attention  & \texttt{y = 0.5 * (z := A @ V) + \detached{[}0.5 * z\detached{].detach()}} \\

            \bottomrule
        \end{tabular}

    \end{tcolorbox}
\end{table}

\section{Results per Task}
\label{fr}
This section details the accuracy scores across all 47 tasks, accounting for every permutation of the two degrees of freedom.

\begin{figure}[ht]
  %\vskip 0.2in
  \begin{center}
    \centerline{\includegraphics[width=0.6\columnwidth]{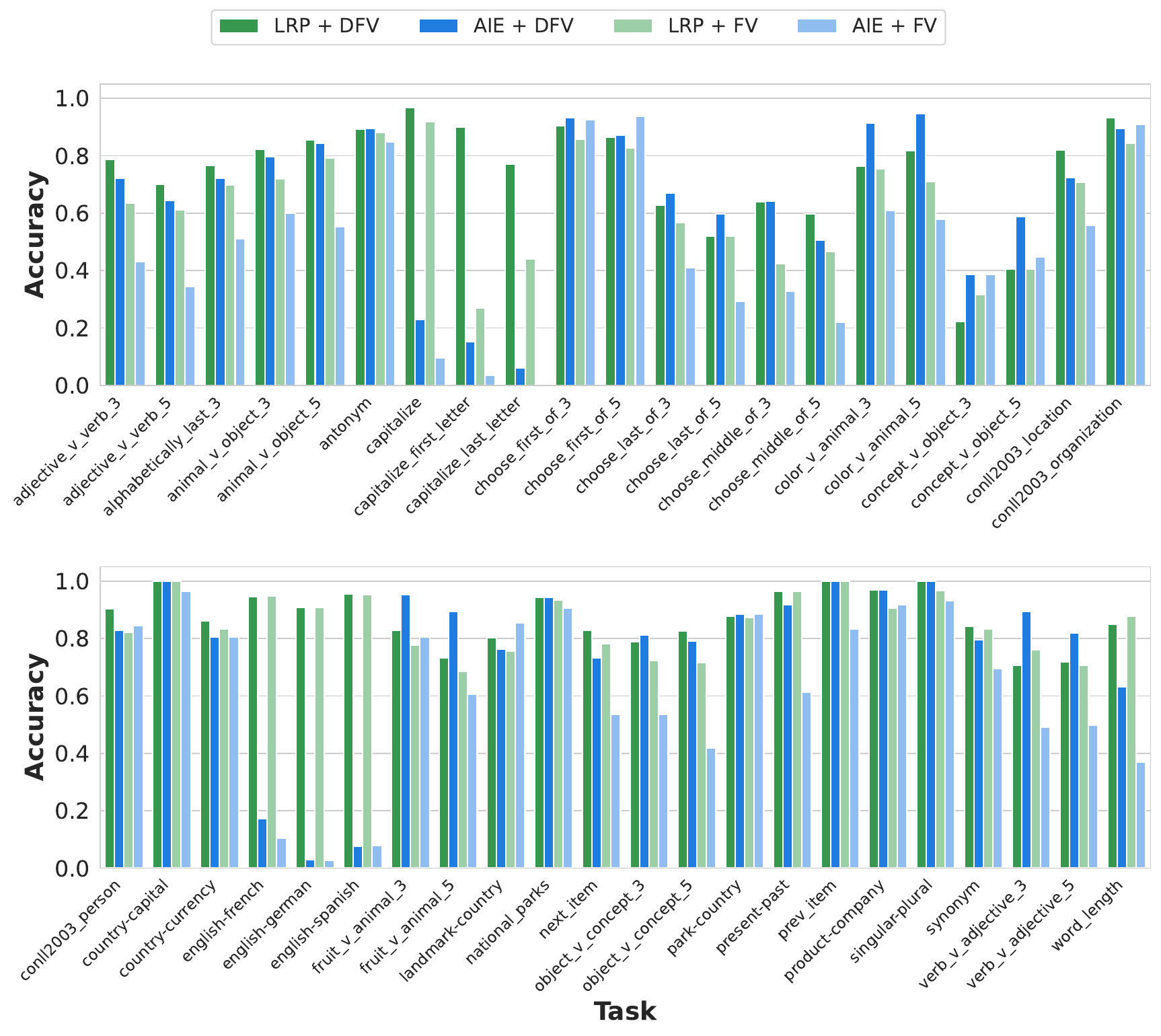}}
    \caption{Accuracies per task with global heads on Llama-3.2-3B.
    }  %TODO add caption
    % TODO higher resulution
    % TODO y-axis to 1.05
    \label{acc_global}
  \end{center}
\end{figure}

\begin{figure}[ht]
  %\vskip 0.2in
  \begin{center}
    \centerline{\includegraphics[width=0.6\columnwidth]{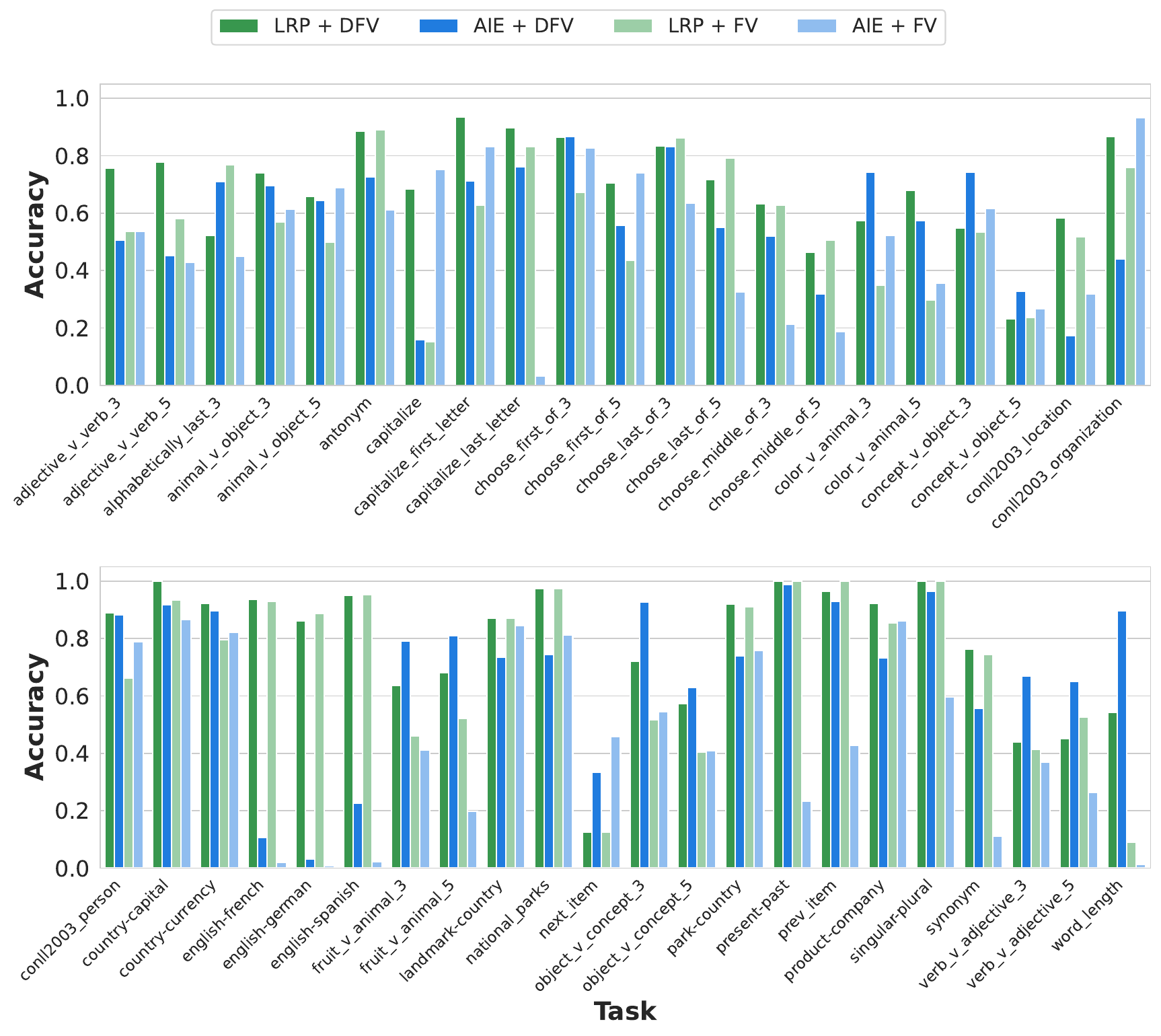}}
    \caption{Accuracies per task with global heads on Llama-3.1-8B.
    }  %TODO add caption
    % TODO higher resulution
    % TODO y-axis to 1.05
  \end{center}
\end{figure}

\begin{figure}[ht]
  %\vskip 0.2in
  \begin{center}
    \centerline{\includegraphics[width=0.6\columnwidth]{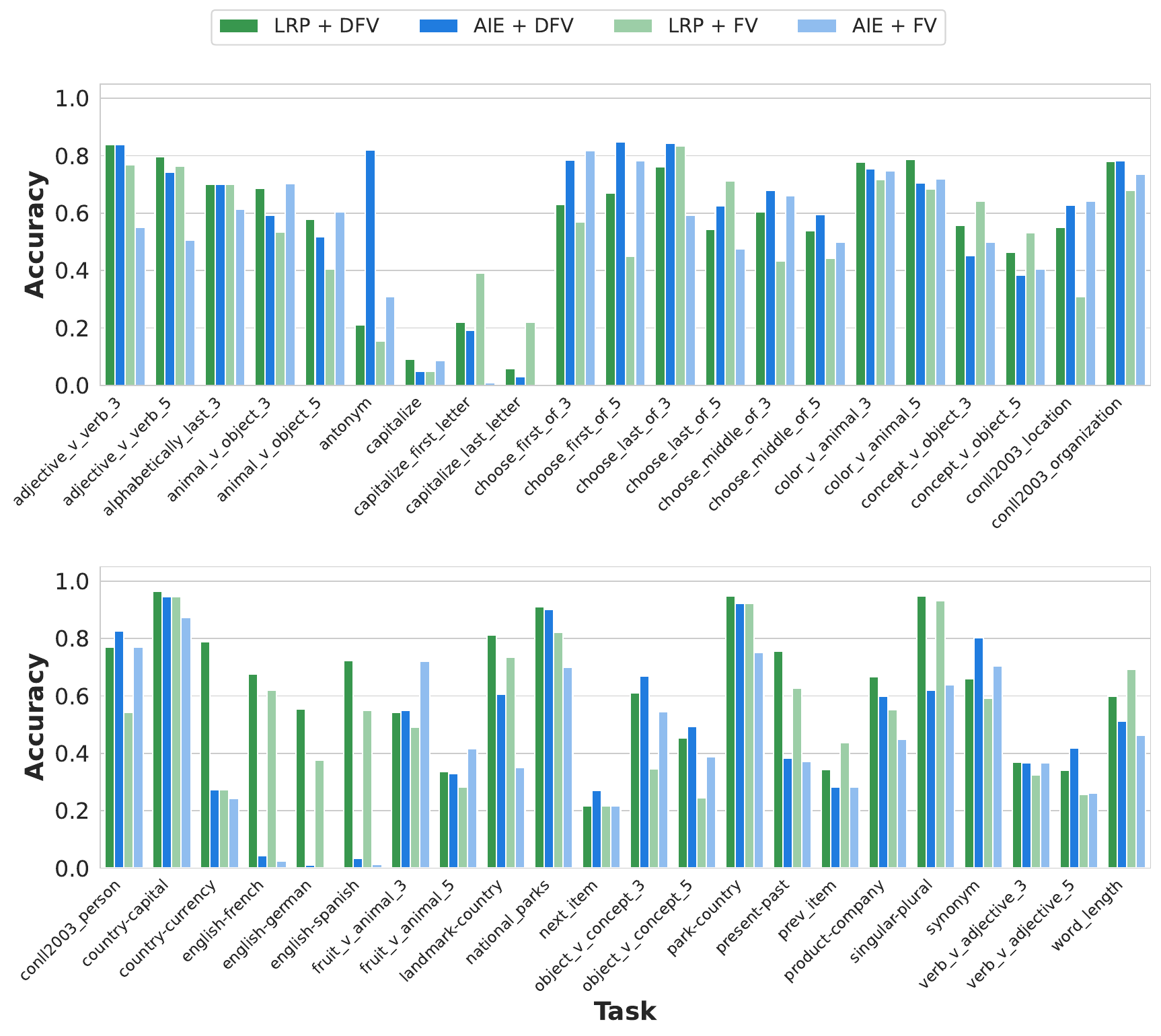}}
    \caption{Accuracies per task with global heads on Qwen3-4B.
    }  %TODO add caption
    % TODO higher resulution
    % TODO y-axis to 1.05
  \end{center}
\end{figure}

\FloatBarrier

\section{Results on Per-Task Heads }
\begin{figure}[ht]
  %\vskip 0.2in
  \begin{center}
    \centerline{\includegraphics[width=0.55\columnwidth]{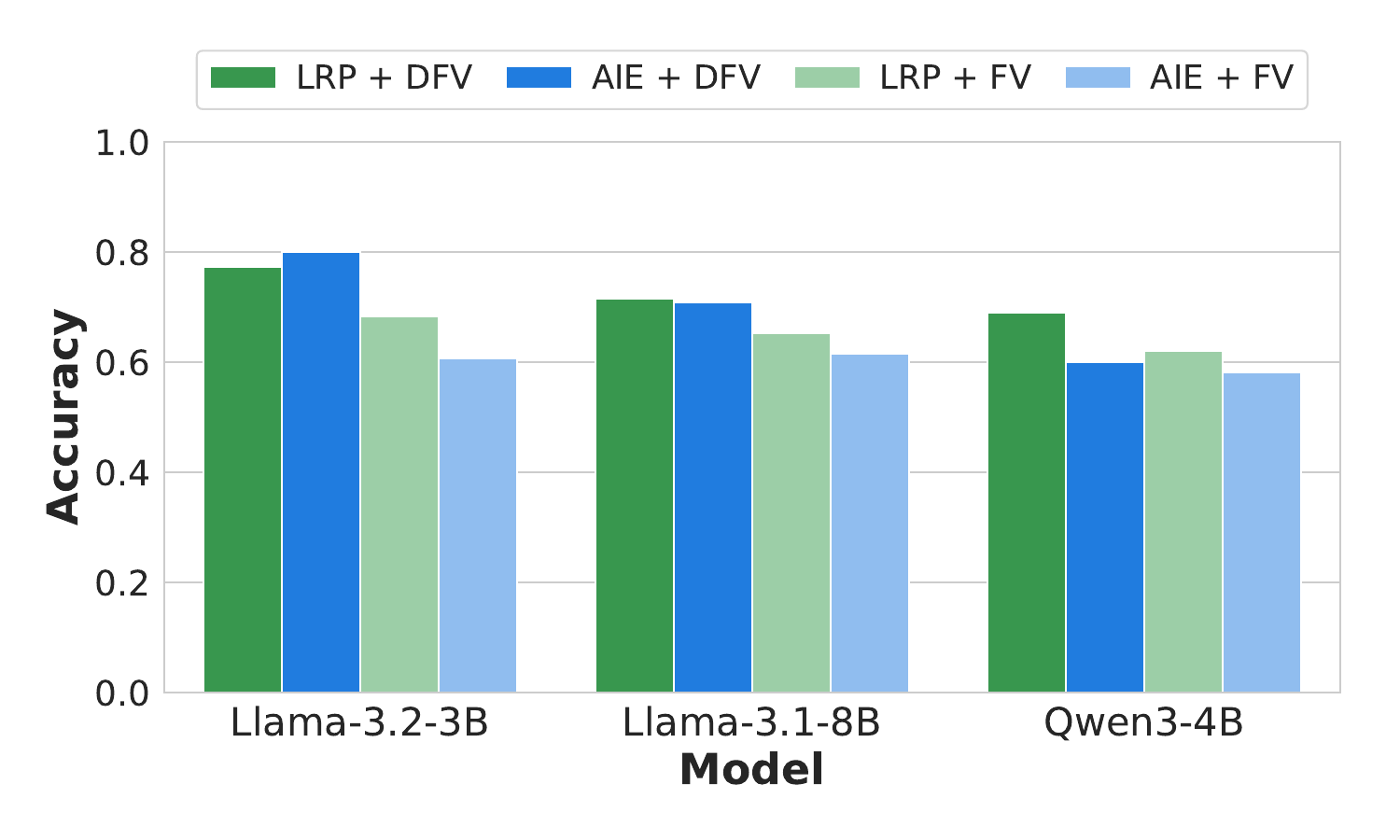}}
    \caption{ Average accuracies over the all tasks with per-task heads.
    }  %TODO add caption
    % TODO higher resulution
    \label{fig:average_acc_per_task1}
  \end{center}
\end{figure}

\begin{figure}[ht]
  %\vskip 0.2in
  \begin{center}
    \centerline{\includegraphics[width=0.55\columnwidth]{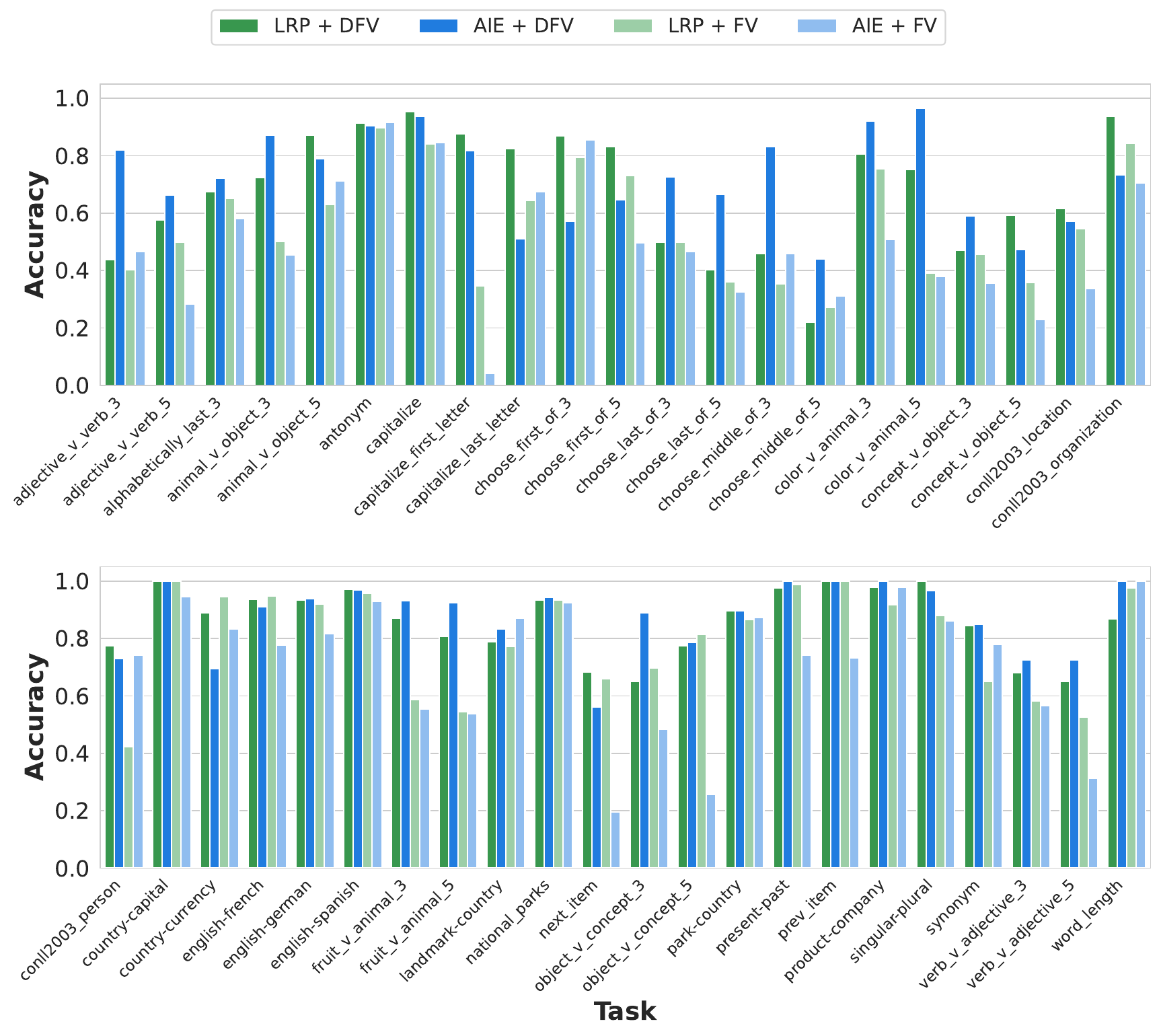}}
    \caption{Accuracies per task with per-task heads on Llama-3.2-3B.
    }  %TODO add caption
    % TODO higher resulution
    \label{fig:acc_pt_llama3b}
  \end{center}
\end{figure}

\begin{figure}[ht]
  %\vskip 0.2in
  \begin{center}
    \centerline{\includegraphics[width=0.55\columnwidth]{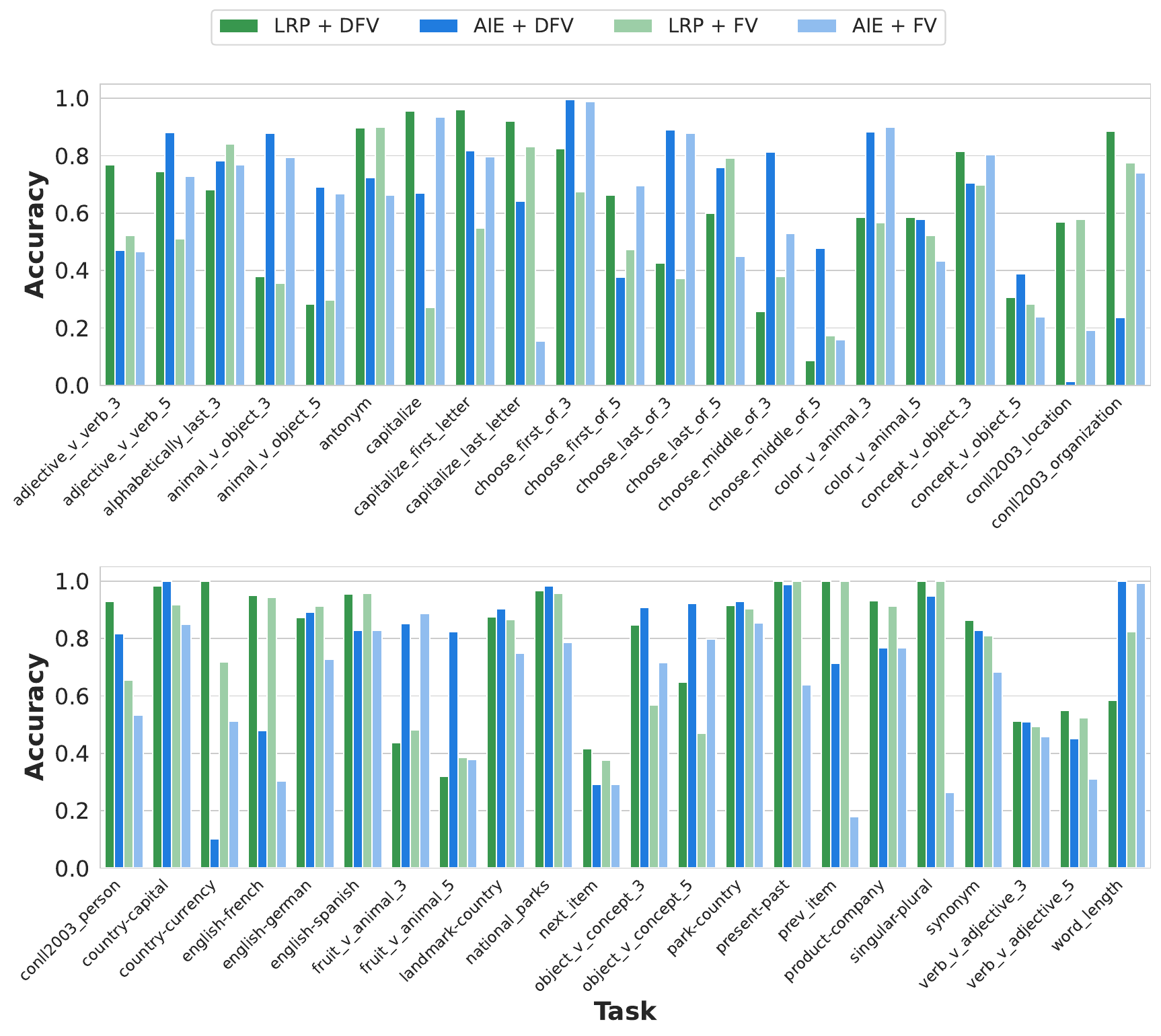}}
    \caption{Accuracies per task with per-task heads on Llama-3.1-8B.
    }  %TODO add caption
    % TODO higher resulution
    \label{fig:acc_pt_llama8b}
  \end{center}
\end{figure}

\begin{figure}[ht]
  %\vskip 0.2in
  \begin{center}
    \centerline{\includegraphics[width=0.55\columnwidth]{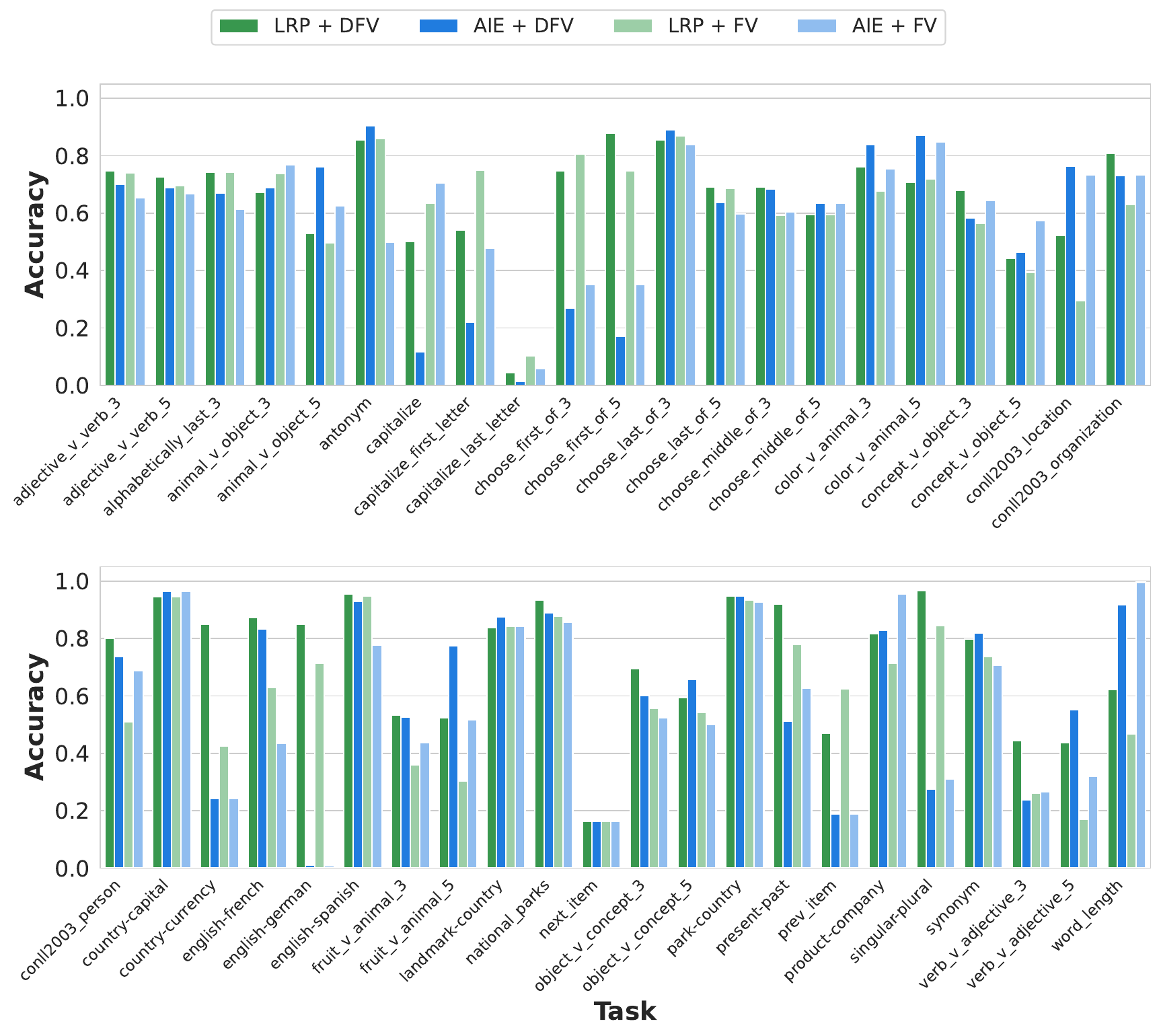}}
    \caption{Accuracies per task with per-task heads on Qwen3-4B.
    }  %TODO add caption
    % TODO higher resulution
    \label{fig:acc_pt_qwen4b}
  \end{center}
\end{figure}

%\todo{maybe make the figures smaller so they fit on the same page}
\label{pth}
We now analyze performance when using separate sets of attention heads per task. Figure~\ref{fig:average_acc_per_task1} reports accuracy averaged across models, with per-task breakdowns for each model shown in Figures~\ref{fig:acc_pt_llama3b} - \ref{fig:acc_pt_qwen4b}.

\newpage
\FloatBarrier
\section{Positions of extracted heads}

We show the positions of the extracted heads for Llama-3.1-8B and Qwen3-4B.

\begin{figure}[ht]
  %\vskip 0.2in
  \begin{center}
    \centerline{\includegraphics[width=0.4\columnwidth]{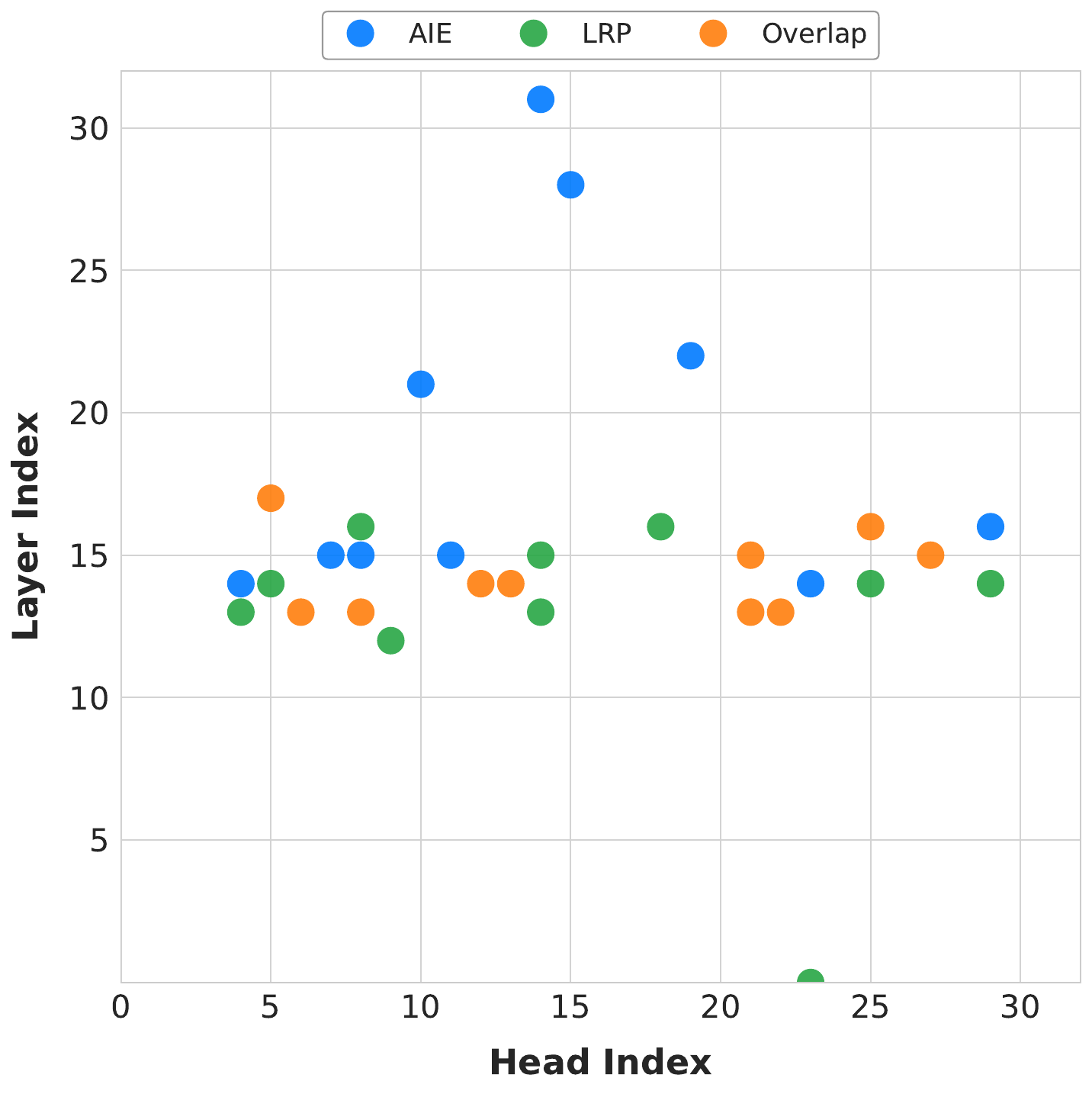}}
    \caption{Position of extracted heads within Llama-3.1-8B.
    } 

    \end{center}
\end{figure}

\begin{figure}[ht]
  %\vskip 0.2in
  \begin{center}
    \centerline{\includegraphics[width=0.4\columnwidth]{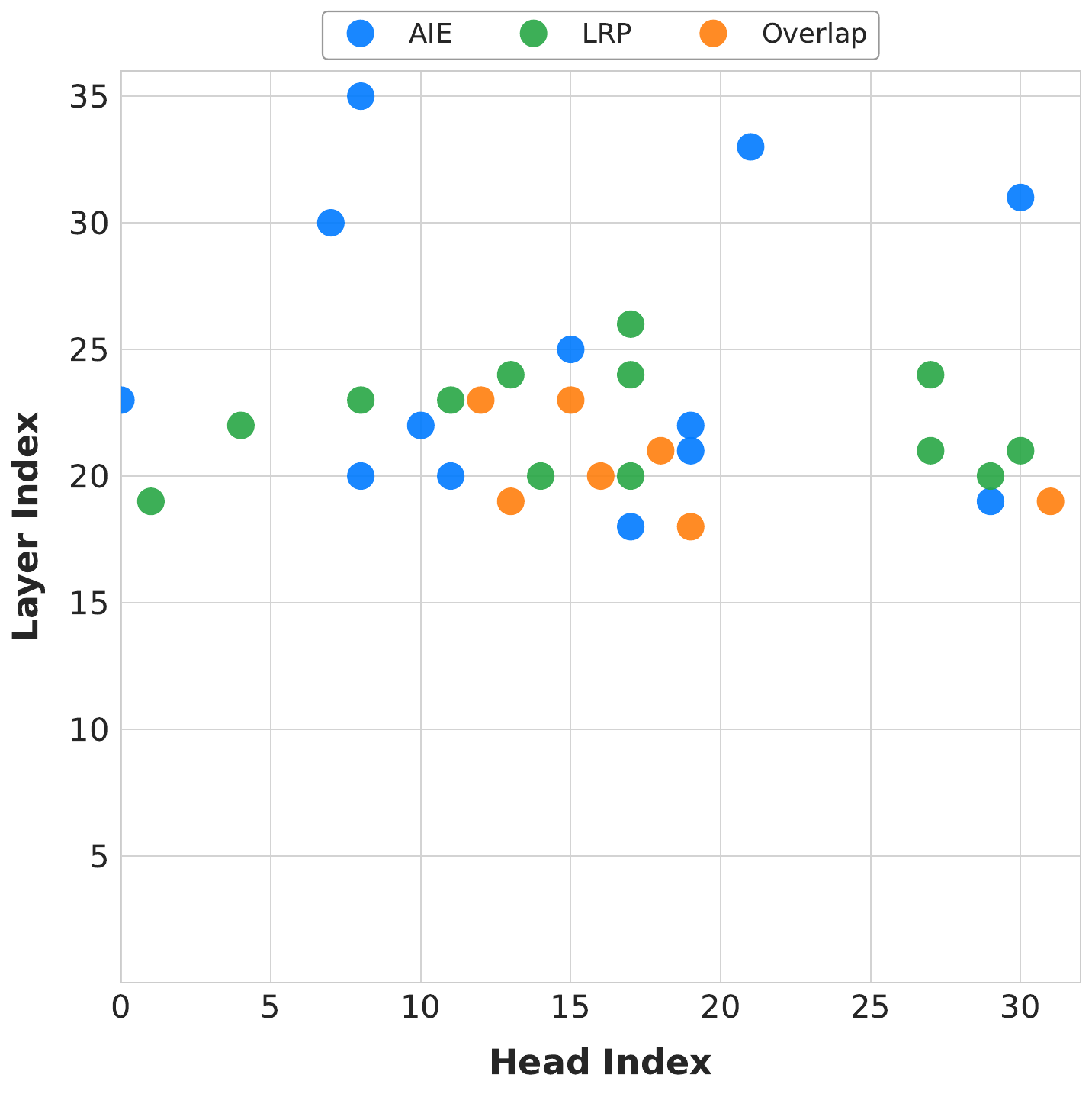}}
    \caption{Position of extracted heads within Qwen3-4B.
    } 

    \end{center}
\end{figure}

\newpage
\FloatBarrier
\section{Optimal Layer Analysis}
\label{ola}

This section shows the results of the search for the optimal application layer for each model, which were used to evaluate the averaged FV method. 

\begin{figure}[ht]
  %\vskip 0.2in
  \begin{center}
    \centerline{\includegraphics[width=0.7\columnwidth]{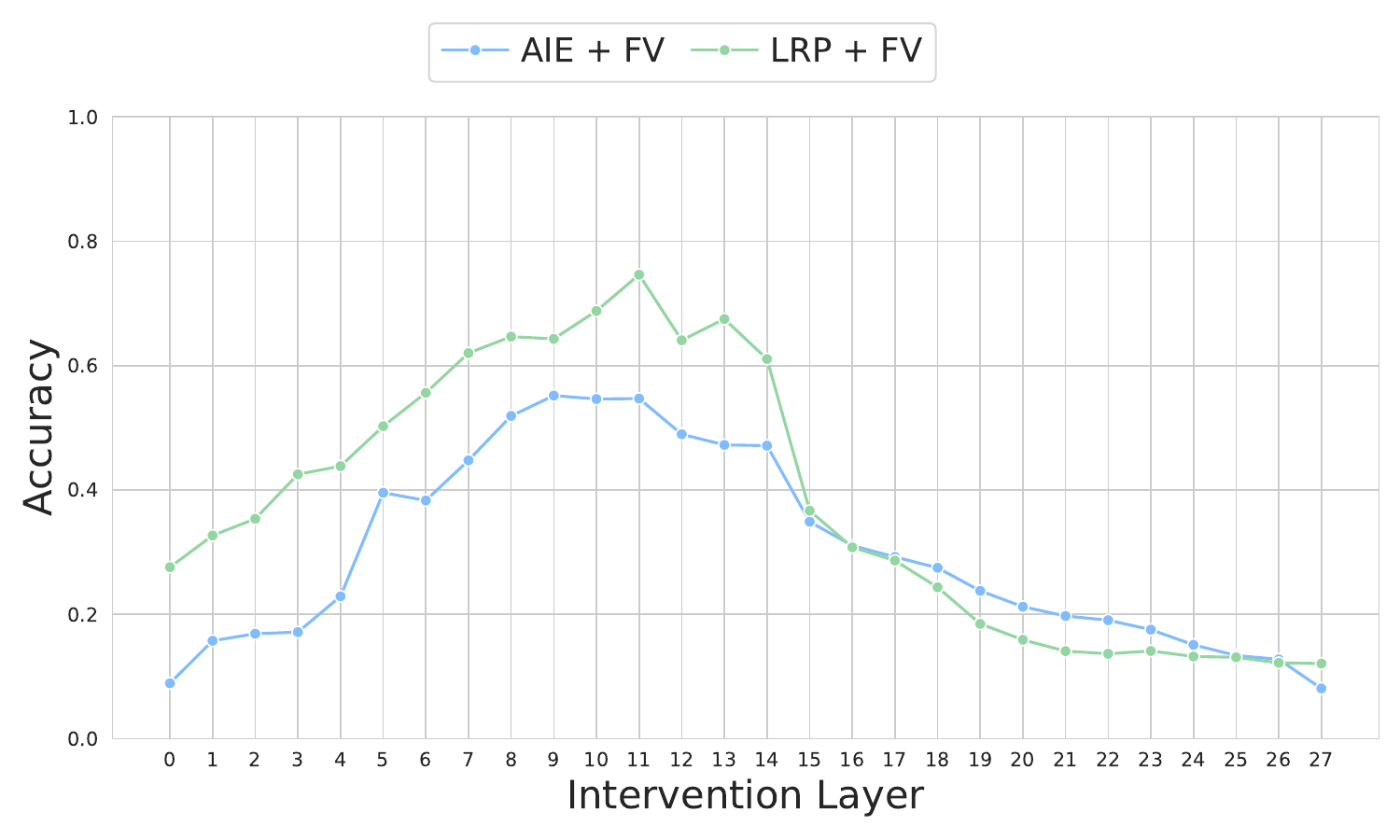}}
    \caption{Accuracies for injecting FVs at different layers for Llama3.2-3B.
    }  %TODO add caption
    % TODO higher resulution
  \end{center}
\end{figure}

\begin{figure}[ht]
  %\vskip 0.2in
  \begin{center}
    \centerline{\includegraphics[width=0.7\columnwidth]{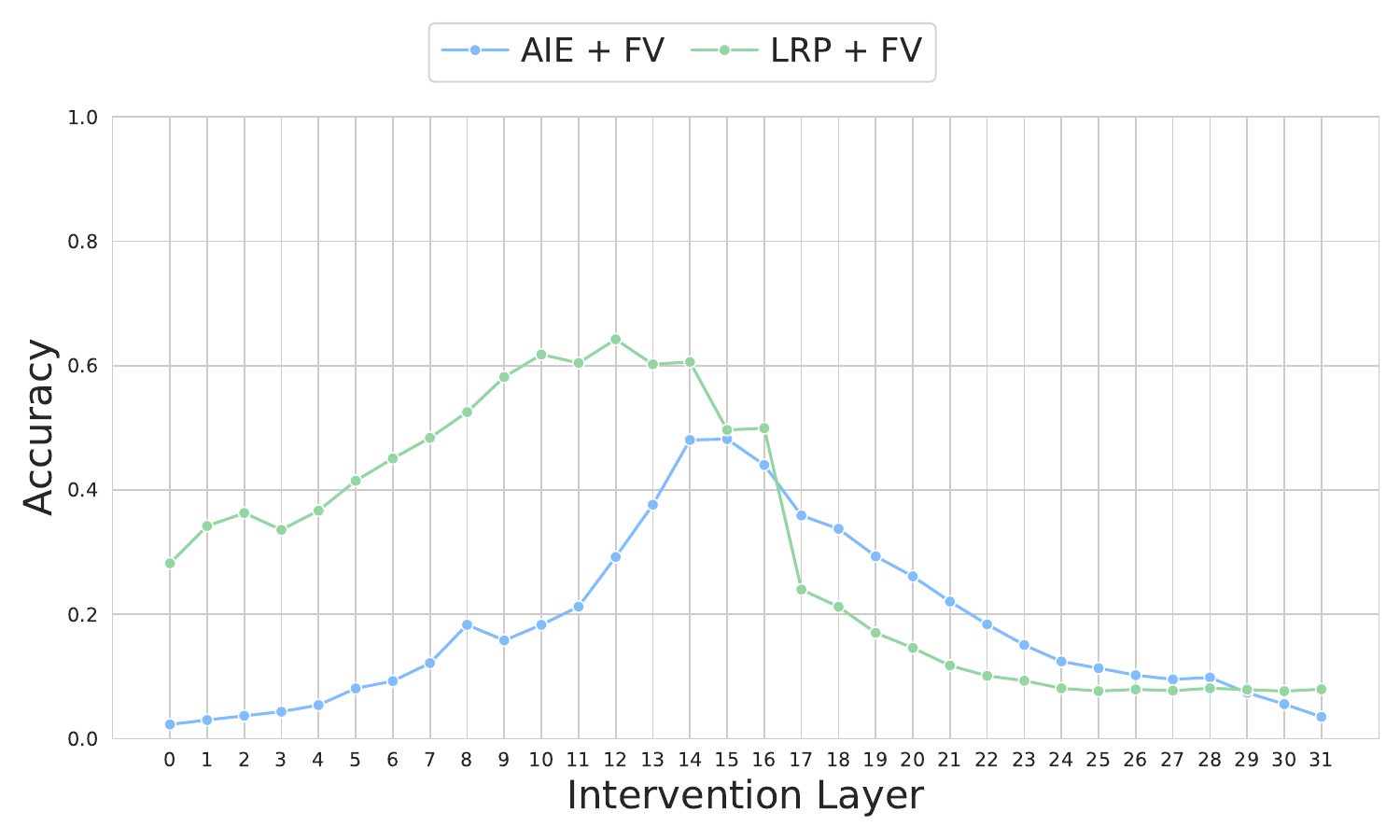}}
    \caption{Accuracies for injecting FVs at different layers for Llama3.1-8B.
    }  %TODO add caption
    % TODO higher resulution
  \end{center}
\end{figure}

\begin{figure}[ht]
  %\vskip 0.2in
  \begin{center}
    \centerline{\includegraphics[width=0.7\columnwidth]{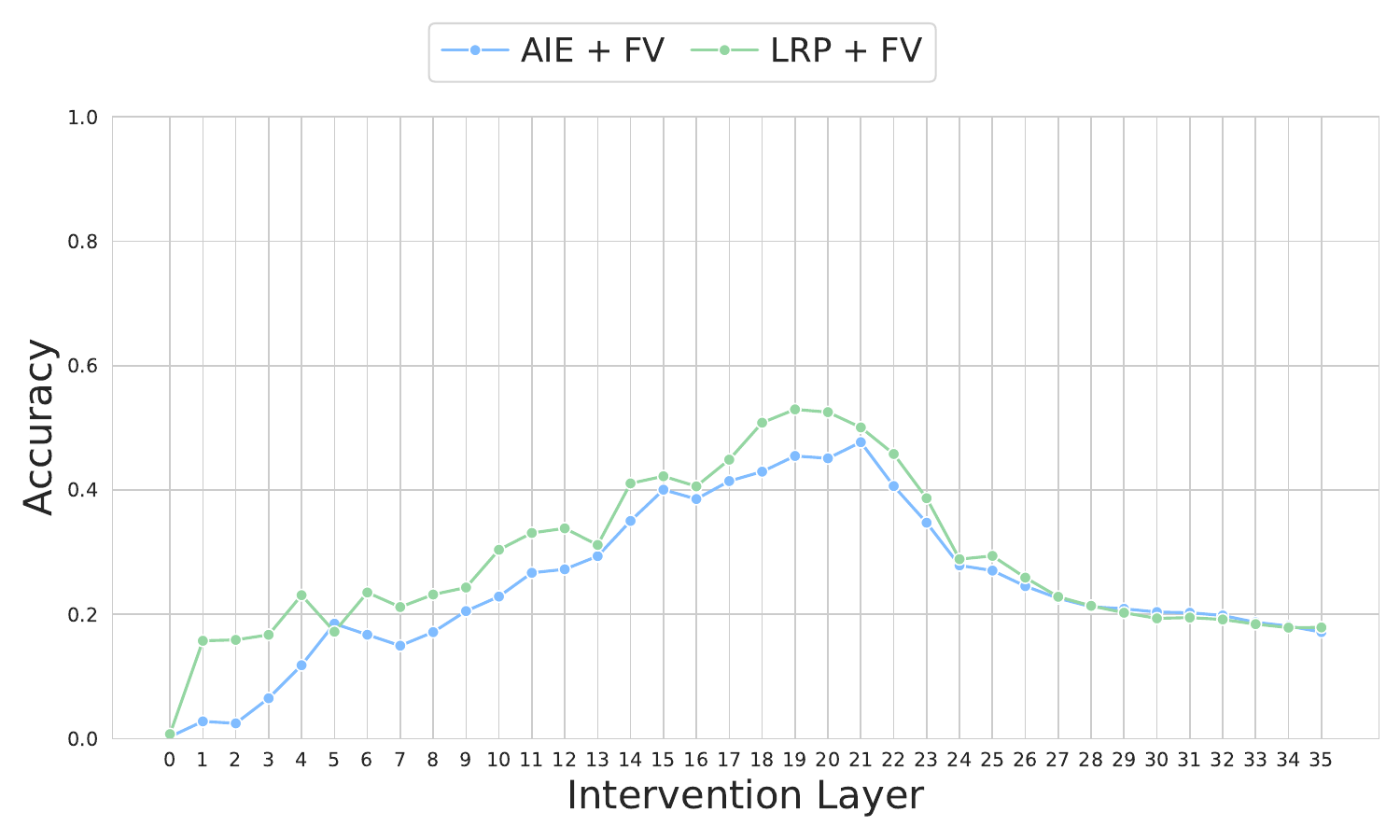}}
    \caption{Accuracies for injecting FVs at different layers for Qwen3-4B.
    }  %TODO add caption
    % TODO higher resulution
  \end{center}
\end{figure}

\FloatBarrier

\section{Exploring larger values for $K$ }
\label{numheads}
\begin{figure}[ht]
  %\vskip 0.2in
  \begin{center}
    \centerline{\includegraphics[width=0.7\columnwidth]{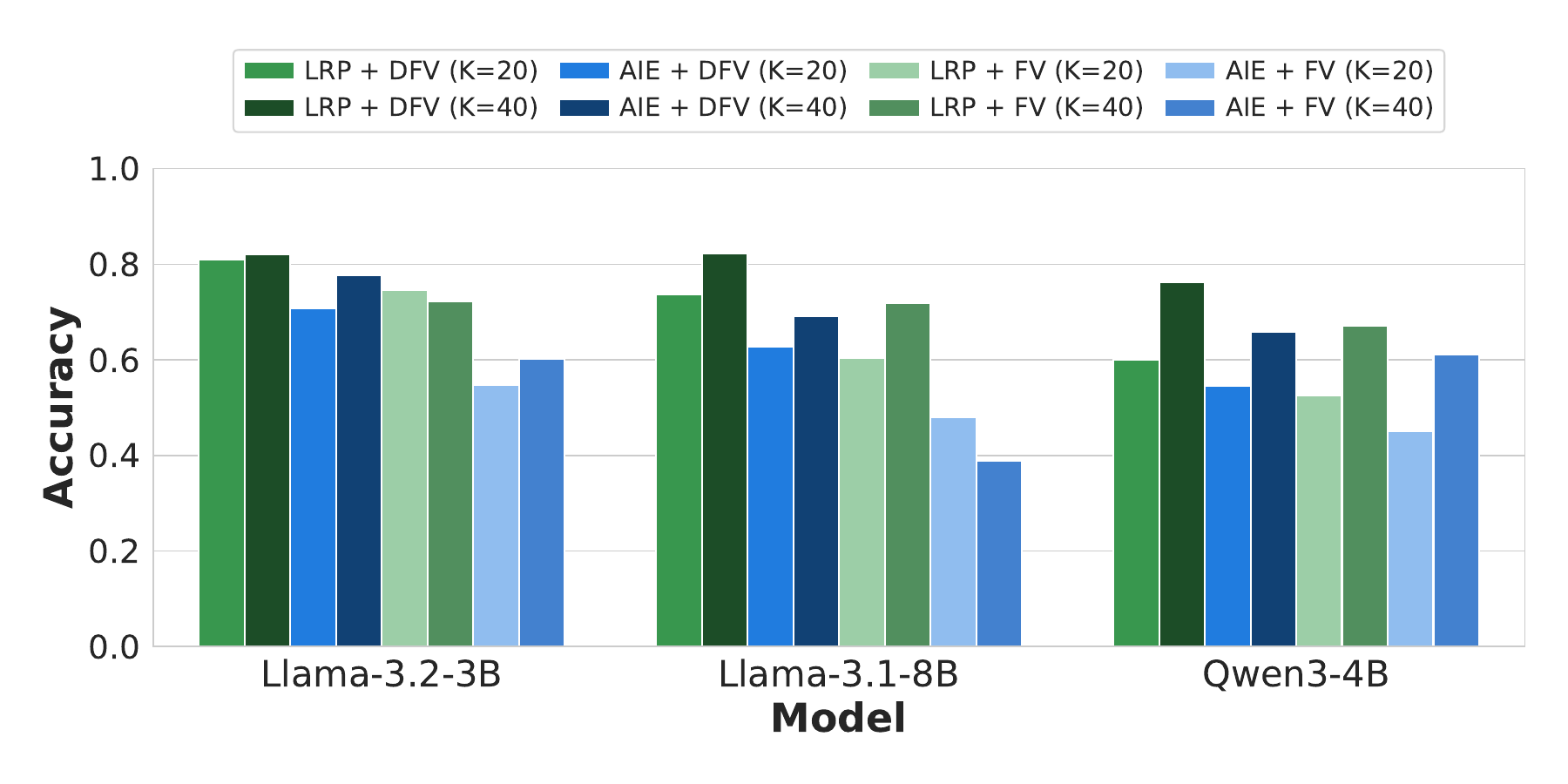}}
    \caption{Conparison of FV patching with $K=20$ and $K=40$.
    }  %TODO add caption
    \label{kheads}
    % TODO higher resulution
  \end{center}
\end{figure}

% TODO llama 8b is worse because we inject less heads relatively
Davidson et~al. \yrcite{davidson2025different} choose $K=20$ for patching FVs, therefore we choose to adapt this. Additionally, we evaluated all models on $K=40$ and show the result in Figure \ref{kheads}. 

\FloatBarrier

\section{List of Tasks}
\label{lot}

We include the list of tasks that were either used or omitted.

\begin{longtable}{@{}p{0.32\textwidth}p{0.28\textwidth}p{0.32\textwidth}@{}}
\caption{Complete list of tasks, filtered out ones are annotated}
\label{tab:tasks}\\
\toprule
\textbf{Task} & \textbf{Citation} & \textbf{Remark} \\
\midrule
\endfirsthead
 
\multicolumn{3}{c}{\tablename\ \thetable{} -- continued from previous page} \\
\toprule
\textbf{Task} & \textbf{Citation} & \textbf{Remark} \\
\midrule
\endhead
 
\midrule \multicolumn{3}{r}{\textit{Continued on next page}} \\
\endfoot
 
\bottomrule
\endlastfoot
 
adjective\_v\_verb\_3       & \citet{todd2023function} & \\ 
adjective\_v\_verb\_5       & \citet{todd2023function} & \\ 
alphabetically\_first\_3    & \citet{todd2023function} & omitted for ill-defined intensional frame  \\ 
alphabetically\_first\_5    & \citet{todd2023function} & omitted for ill-defined intensional frame  \\ 
alphabetically\_last\_3     & \citet{todd2023function} &  \\
alphabetically\_last\_5     & \citet{todd2023function} & omitted for ill-defined intensional frame  \\
animal\_v\_object\_3        & \citet{todd2023function} & \\
animal\_v\_object\_5        & \citet{todd2023function} & \\
antonym                     & \citet{nguyen2017} & \\
capitalize                  & \citet{todd2023function} & \\
capitalize\_first\_letter   & \citet{todd2023function} & \\
capitalize\_last\_letter    & \citet{yin2025attentionheadsmatterincontext}    & \\
capitalize\_second\_letter  & \citet{yin2025attentionheadsmatterincontext}    & omitted for misssing req. performance \\
choose\_first\_of\_3        & \citet{todd2023function} & \\
choose\_first\_of\_5        & \citet{todd2023function} & \\
choose\_last\_of\_3         & \citet{todd2023function} & \\
choose\_last\_of\_5         & \citet{todd2023function} & \\
choose\_middle\_of\_3       & \citet{todd2023function} & \\
choose\_middle\_of\_5       & \citet{todd2023function} & \\
color\_v\_animal\_3         & \citet{todd2023function} & \\
color\_v\_animal\_5         & \citet{todd2023function} & \\
concept\_v\_object\_3       & \citet{todd2023function} & \\
concept\_v\_object\_5       & \citet{todd2023function} & \\
conll2003\_location         & \citet{sang2003introduction} & \\
conll2003\_organization     & \citet{sang2003introduction} & \\
conll2003\_person           & \citet{sang2003introduction} & \\
country-capital             & \citet{todd2023function} & \\
country-currency            & \citet{todd2023function} & \\
english-french              & \citet{lample2018word} & \\
english-german              & \citet{lample2018word} & \\
english-spanish             & \citet{lample2018word} & \\
fruit\_v\_animal\_3         & \citet{todd2023function} & \\
fruit\_v\_animal\_5         & \citet{todd2023function} & \\
landmark-country            & \citet{hernandez2023} & \\
lowercase\_first\_letter    & \citet{todd2023function} & omitted for misssing req. performance \\  
lowercase\_last\_letter     & \citet{todd2023function} & omitted for misssing req. performance \\   
national\_parks             & \citet{todd2023function} & \\
next\_capital\_letter       & \citet{todd2023function} & omitted for misssing req. performance \\  
next\_item                  & \citet{todd2023function} & \\
object\_v\_concept\_3       & \citet{todd2023function} & \\
object\_v\_concept\_5       & \citet{todd2023function} & \\
park-country                & \citet{todd2023function} & \\
present-past                & \citet{todd2023function} & x \\  
prev\_item                  & \citet{todd2023function} & \\
product-company             & \citet{hernandez2023} & \\
singular-plural             & \citet{todd2023function} & \\
synonym                     & \citet{nguyen2017} & \\
verb\_v\_adjective\_3       & \citet{todd2023function} & \\
verb\_v\_adjective\_5       & \citet{todd2023function} & \\
word\_length                & \citet{todd2023function} & \\
\end{longtable}
\newpage
\section{Example Prompts and Intensional Frames}
\label{epif}

Below we provide example prompts from 5 tasks with their respective intensional frame underlined

\begin{table}[h]
\caption{Example tasks and prompt, intensional frame underlined.}
\label{tab:example_prompts}
\centering
\renewcommand{\arraystretch}{1.3}
\begin{tabular}{|p{0.25\linewidth}|p{0.65\linewidth}|}
\hline
\textbf{Task} & \textbf{Prompt} \\
\hline
antonym & Provide a word that is the semantic \underline{opposite} \newline Q: inside \newline A: outside \\
\hline
country & Find the \underline{country} where a specific national park is located \newline Q: Wilpattu National Park \newline A: Sri Lanka \\
\hline
prev\_item & Go one position \underline{backward} in the sequence \newline Q: March \newline A: February \\
\hline
fruit\_v\_animal\_5 & Extract the \underline{fruit} word from the given list \newline Q: apricot, toucan, mantis, reindeer, donkey \newline A: apricot \\
\hline
concept\_v\_object\_5 & Find the word that is \underline{not a concrete object} \newline Q: envelope, cheerful, beef, net, stapler \newline A: cheerful \\
\hline
\end{tabular}

\end{table}

\section{Failure Analysis}

In Table \ref{tab:antonym-failures} we list all failed samples steering with LRP-selected heads and DFVs on the \textit{antonym} task dataset and evaluate each of them separately.

% Annotated table of all failed predictions from the antonym task.
% Requires \usepackage{booktabs} for \toprule, \midrule, \bottomrule.
\begin{table}[ht]
    \centering
    \small
        \caption{Qualitative analysis of all 41 failed predictions on the \textit{antonym} task. We don't consider the first two groups genuine task failures as they correspond to valid synonyms of the target or sub-word prefixes that plausibly continue into a valid antonym under multi-token decoding. The remaining 15 cases constitute genuine failures and are further sub-categorised. The \textit{rank} column reports the rank of the target token in the model's output distribution.}
    \label{tab:antonym-failures}
    \begin{tabular}{llllp{4.5cm}}
        \toprule
        \textbf{Input} & \textbf{Target} & \textbf{Predicted} & \textbf{Rank} & \textbf{Annotation} \\
        \midrule
        \multicolumn{5}{l}{\textit{Synonyms of the target (17)}} \\
        \midrule
        goodness    & evil        & bad        & 2  & Synonym of target \\
        doubt       & certainty   & faith      & 2  & Synonym of target \\
        order       & chaos       & disorder   & 3  & Synonym of target \\
        magical     & mundane     & ordinary   & 3  & Synonym of target \\
        rotten      & fresh       & sound      & 2  & Synonym of target \\
        sluggish    & energetic   & brisk      & 8  & Synonym of target \\
        hard        & soft        & easy       & 2  & Antonym in difficulty sense \\
        obsolete    & modern      & current    & 2  & Synonym of target \\
        approval    & disapproval & rejection  & 2  & Synonym of target \\
        recession   & boom        & expansion  & 2  & Synonym of target \\
        subtle      & obvious     & overt      & 2  & Synonym of target \\
        dumb        & smart       & intelligent& 2  & Synonym of target \\
        interesting & boring      & dull       & 3  & Synonym of target \\
        segregated  & integrated  & unified    & 2  & Synonym of target \\
        pleased     & displeased  & unhappy    & 2  & Synonym of target \\
        usual       & unusual     & uncommon   & 2  & Synonym of target \\
        adulthood   & childhood   & youth      & 2  & Synonym of target \\
        \midrule
        \multicolumn{5}{l}{\textit{Sub-word prefixes plausibly continuing into a valid antonym (9)}} \\
        \midrule
        polite      & rude        & imp        & 2  & Likely \textit{impolite} \\
        fit         & unfit       & not        & 2  & Negation prefix \\
        radical     & conservative& non        & 4  & Likely \textit{non-radical} \\
        separation  & union       & un         & 3  & Likely \textit{union}/\textit{unity} \\
        oust        & install     & in         & 27 & Likely \textit{install}/\textit{instate} \\
        agricultural& industrial  & non        & 2  & Likely \textit{non-agricultural} \\
        privacy     & publicity   & public     & 2  & Likely \textit{publicity} \\
        fascism     & democracy   & anti       & 3  & Likely \textit{anti-fascism} \\
        volatile    & stable      & non        & 2  & Likely \textit{non-volatile} \\
        \midrule
        \multicolumn{5}{l}{\textit{Synonyms of the input (6)}} \\
        \midrule
        huge        & tiny        & enormous   & 3  & Synonym of input \\
        lifelong    & temporary   & permanent  & 8  & Synonym of input \\
        impede      & facilitate  & hinder     & 2  & Synonym of input \\
        renowned    & unknown     & infamous   & 3  & Valence-shifted synonym of input \\
        power       & weakness    & energy     & 5  & Synonym of input \\
        epidemic    & endemic     & pandemic   & 2  & Magnitude variant of input \\
        \midrule
        \multicolumn{5}{l}{\textit{Verbatim copies of the input (3)}} \\
        \midrule
        supreme     & inferior    & supreme    & 2  & Input copied \\
        necessity   & luxury      & necessity  & 20 & Input copied \\
        preferable  & undesirable & preferable & 4  & Input copied \\
        \midrule
        \multicolumn{5}{l}{\textit{Semantically related but non-antonymic (6)}} \\
        \midrule
        copy        & original    & paste      & 2  & Complement, not opposite \\
        futures     & pasts       & options    & 3  & Wrong sense of input \\
        rocky       & smooth      & still      & 4  & Antonym of different sense \\
        mainland    & island      & overseas   & 2  & Adjacent, not antonymic \\
        erase       & write       & keep       & 11 & Loosely related \\
        elaborate   & simple      & to         & 17 & Function word \\
        \bottomrule
    \end{tabular}

\end{table}